\documentclass[conference]{IEEEtran}

\IEEEoverridecommandlockouts
\usepackage{cite}
\usepackage{amsmath,amssymb,amsfonts}
\usepackage{graphicx}
\usepackage{textcomp}
\usepackage{xcolor}
\usepackage{subcaption} 
\usepackage{multirow}
\usepackage{CJKutf8} 
\usepackage{cleveref} 
\usepackage{algorithm}
\usepackage{algpseudocode}
\usepackage[export]{adjustbox}
\usepackage{changepage}
\usepackage{float}
\usepackage[normalem]{ulem}

\def\BibTeX{{\rm B\kern-.05em{\sc i\kern-.025em b}\kern-.08em
    T\kern-.1667em\lower.7ex\hbox{E}\kern-.125emX}}

\begin{document}

%%%%%%%%%%%%%%%%%%%%%%%%%%%%%%%%%%
%
%    TITLE
%
%%%%%%%%%%%%%%%%%%%%%%%%%%%%%%%%%%%
\title{A Study of Fitness Landscapes for Neuroevolution}

%%%%%%%%%%%%%%%%%%%%%%%%%%%%%%%%%%
%
%    AUTHORS
%
%%%%%%%%%%%%%%%%%%%%%%%%%%%%%%%%%%%
\author{\IEEEauthorblockN{Nuno M. Rodrigues}
\IEEEauthorblockA{LASIGE, \\ Departamento de Inform\'atica,\\ Faculdade de Ci\^encias,\\ Universidade de Lisboa,\\ 1749-016 Lisboa, Portugal\\
nmrodrigues@fc.ul.pt}
\and
\IEEEauthorblockN{Sara Silva}
\IEEEauthorblockA{LASIGE, \\ Departamento de Inform\'atica,\\ Faculdade de Ci\^encias,\\ Universidade de Lisboa, \\ 1749-016 Lisboa, Portugal\\
sara@fc.ul.pt}
\and
\IEEEauthorblockN{Leonardo Vanneschi}
\IEEEauthorblockA{NOVA Information Management \\ School (NOVA IMS),\\ Universidade Nova de Lisboa, \\ Campus de Campolide, \\ 1070-312 Lisboa, Portugal\\
lvanneschi@novaims.unl.pt}
}

%%%%%%%%%%%%%%%%%%%%%%%%%%%%%%%%%%
%
%    Maketitle
%
%%%%%%%%%%%%%%%%%%%%%%%%%%%%%%%%%%%
\maketitle

%%%%%%%%%%%%%%%%%%%%%%%%%%%%%%%%%%%%%%
%
%    ABSTRACT
%
%%%%%%%%%%%%%%%%%%%%%%%%%%%%%%%%%%%%%%
\begin{abstract}
Fitness landscapes are a useful concept  to study the dynamics of meta-heuristics.
In the last two decades, they have been applied with success to estimate the optimization power of several types of evolutionary algorithms, including genetic algorithms and genetic programming.
However, so far they have never been used to study the performance of machine learning
algorithms on unseen data, and they have never been applied to neuroevolution. 
This paper aims at filling both these  gaps, applying for the first time fitness landscapes to neuroevolution and using them to infer useful information about the predictive ability of the method.
More specifically, we use a grammar-based approach to generate  convolutional neural networks, and we study the dynamics of three different mutations
to evolve them.
To characterize fitness landscapes, we study  autocorrelation and entropic measure of ruggedness.
The results show that these measures are appropriate for estimating both the optimization power 
and the generalization ability of the considered neuroevolution configurations.
\end{abstract}

%%%%%%%%%%%%%%%%%%%%%%%%%%%%%%%%%%%%%
%
%    KEYWORDS
%
%%%%%%%%%%%%%%%%%%%%%%%%%%%%%%%%%%%%

%%%%%%%%%%%%%%%%%%%%%%%%%%%%%
%
%   Introduction
%
%%%%%%%%%%%%%%%%%%%%%%%%%%%%%
\section{Introduction}
\label{intro}

Fitness landscapes~(FLs)~\cite{Wright1932, Stadler2002}  have been studied for many years to 
characterize the dynamics of meta-heuristics in optimization.
In particular, several measures have been introduced to capture some important characteristics 
of FLs, that can give useful information about the difficulty of optimization problems.
Among those measures, autocorrelation~\cite{Weinberger1990} and entropic measure of ruggedness~(EMR)~\cite{Vassilev:phd,Vassilev2000,Vassilev2003} have been 
intensively studied, and they revealed useful indicators of the ruggedness of the~FLs
induced by several variants of local search meta-heuristics and evolutionary algorithms~(EAs)~\cite{vereltut}.
However, to the best of our knowledge, no measure related to~FLs
has ever been applied so far to study the performance 
of machine learning~(ML) algorithms on unseen data,
and none of those measures has ever been used to characterize the functioning
of neuroevolution.
In this work, for the first time, we adapt the well-known definitions
of autocorrelation and~EMR
to neuroevolution, and we use those measures not only to study the optimization effectiveness
of various configurations of the method, but also to characterize their 
performance on unseen data.

Neuroevolution is a branch of evolutionary computation that has been used for almost three decades, 
with application in multiple areas such as supervised 
classification tasks~\cite{BALDOMINOS} and agent building~\cite{neat}.
In neuroevolution, an~EA is used to evolve, for instance, 
weights, 
topologies and/or hyper-parameters of artificial neural networks. 
In this study, we focus on the evolution of convolutional neural networks
(CNNs), not only because they are arguably one of
the most popular deep neural network architectures, but also
because they have a vast amount of tunable parameters, which
makes CNNs appropriate to test the capabilities of neuroevolution.
We use our own
%The study will be carried on using a  
grammar-based neuroevolution
approach, inspired by existing systems, also introduced here.
%is presented here for the first time.

This work is aimed at testing the reliability of autocorrelation and~EMR
in predicting the performance of neuroevolution of~CNNs on training and 
unseen data.
For achieving this task, we consider three different types of mutations and four 
different multiclass classification problems, with different degrees of difficulty.
For each type of mutation, and for each one of the studied problems,
we calculate the value of these measures and we
compare them to the results obtained by actual simulations of our 
neuroevolution system, to test the reliability of the measures.
We consider this work as the first proof of concept 
in a wider study, aimed at establishing the use of FL~measures as indicators
to characterize neuroevolution of~CNNs.
If successful, this study will be extremely impactful.
In fact,
CNNs usually have a slow learning phase,
which makes neuroevolution a very intensive computational process, since it requires the evaluation of several~CNNs in each generation.
%and consequently, neuroevolution, that
%works by evaluating several~CNNs at the same time,
%is computationally very intensive.
For this reason, the task of executing simulations to choose
among different types of genetic operators, and/or among several possible parameter settings,
is generally prohibitively expensive.
On the other hand, the calculation of measures such as the autocorrelation
and the~EMR is much faster.
So, these measures could help us find appropriate neuroevolution configurations 
much more efficiently.

The paper is organized as follows: 
in Section~\ref{fl}, we introduce the concept of~FL and the used measures.
Section~\ref{ne} introduces neuroevolution and presents our 
grammar-based approach to evolve CNNs.
In Section~\ref{estud}, we present our experimental study,
first discussing the used test problems and the experimental settings,
and then presenting the obtained results.
Finally, Section~\ref{concl} concludes the paper and suggests ideas
for future research.

%%%%%%%%%%%%%%%%%%%%%%%%%%%%%
%
%   Fitness Landscapes
%
%%%%%%%%%%%%%%%%%%%%%%%%%%%%%
\section{Fitness Landscapes}
\label{fl}

Using a landscape metaphor to gain insight about the workings of a
complex system originates with the work of Wright on genetics~\cite{Wright1932}.
Probably, the simplest definition of~FL 
is the following one: a~FL is a plot where the
points in the horizontal direction represent the different individual genotypes
in a search space and the points in the vertical direction
represent the fitness of each one of these individuals~\cite{langdon-poli-02}.
If genotypes can be visualized in two dimensions, the plot can be
seen as a three-dimensional ``map'', which may contain peaks and valleys.
The task of finding the best solution to the problem is equivalent
to finding the highest peak (for maximization problems) or the lowest valley (for minimization).
The problem solver is seen as a short-sighted explorer searching for those optimal spots.
Crucial to the concept of~FL is that solutions should 
be arranged 
in a way that is consistent with a given neighborhood relationship. 
Indeed, a FL is completely identified by the triple
$(S, f, N)$,
where~$S$ is the set of all admissible solutions (the search space), 
$f$ is the fitness function, and~$N$ is the neighborhood.
Generally, the neighborhood~$N$ should have a relationship with the transformation (mutation) operator used to explore the search space. 
A typical example is to consider as neighbors two solutions~$a$ and~$b$ if and only if 
$b$ can be obtained by applying mutation to~$a$.

The~FL metaphor can be helpful to understand the
difficulty of a problem, i.e., the ability of a searcher to find
the optimal solution for that problem.
%
%
% For example, imagine a very smooth and regular landscape 
% with a single hill top.
% This is the typical fitness landscape of an easy problem, where all
% searching strategies are able to find the top
% of the hill in a straighforward manner.
% The opposite is true for a very rugged landscape, with many hills which are not
% as high as the best one.
% In this case, searching strategies will probably get stuck in local optima.
%
%
However, in practical situations, FLs are impossible to visualize, 
both because of the vast size of the search space~$S$ and because of the 
multi-dimensionality of the neighborhood~$N$.
For this reason, researchers have introduced a set of mathematical measures, 
able to capture some characteristics of FLs and express them with 
a single number~\cite{thesis}. 
Although none of these measures is capable of expressing completely the vast amount of information 
that characterize a FL, some of them revealed to be reliable indicators of the difficulty of problems, for instance:
autocorrelation~\cite{Weinberger1990}, 
entropic measure of ruggedness~(EMR)~\cite{Vassilev:phd,Vassilev2000,Vassilev2003},
density of states~\cite{Rose},
fitness-distance correlation~\cite{jones95a,thesis},
length of adaptive walks~\cite{Stadler2002},
basins of attraction size~\cite{oc10},
plus various measures based of the concepts of fitness clouds~\cite{thesis} and 
local optima networks~\cite{Ochoa}.
In this paper, we have decided to investigate autocorrelation and EMR
because, among the previously mentioned measures, they are probably
the most simple ones, and also the ones that can be calculated more efficiently.
For this reason, they seem appropriate for studying
%for the first time
a computationally intensive method such as neuroevolution of~CNNs.
Autocorrelation and EMR are presented next.
%in the continuation of this section.

%%%%%%%%%%%%%%%%%%%%%%%%%%%%%
%
%   Autocorrelation
%
%%%%%%%%%%%%%%%%%%%%%%%%%%%%%
\subsection{Autocorrelation}
\label{autocorr}

The autocorrelation coefficient is used to measure the ruggedness of a~FL~\cite{Weinberger1990}.
It is applied over a series of fitness values, determined by a {\it walk} on the landscape. 
A walk on a~FL is a sequence of solutions
$(s_{0}, s_{1},..., s_{n})$,
such that for each $t = 1,2,...,n$, $s_{t}$ 
is a neighbor of~$s_{t-1}$ or, in other words,
$s_{t}$ is obtained by applying mutation to~$s_{t-1}$. 
%
%
%
% Autocorrelation with a {\it step}~$k$ is defined as:
%
% \footnotesize
% \begin{equation*}
%    \rho(k) = \frac{E[f(s_{t})f(s_{t+k})]-E[f(s_{t})]E[f(s_{t+k})]}
%                   {\sqrt{E[f(s_{t})^2]-[E[f(s_{t})]]^2}-\sqrt{E[f(s_{t+k})^2]-[E[f(s_{t+k})]]^2}} ,
% \end{equation*}
%\normalsize
%
% \noindent
% where $E[f(s_{t})]$ is the expected value of $f(s_{t})$ and $k$
% indicates how far apart each solution is from the next one considered in the walk.
%
%
%
For walks of a finite length~$n$, autocorrelation  with a {\it step}~$k$ is defined as:

\begin{equation*}
    \hat{\rho}(k) = \frac{\sum_{t=1}^{n-k}(f(s_{t})-\bar{f})(f(s_{t+k})-\bar{f})}
                      {\sqrt{\sum_{t=1}^{n}(f(s_{t})-\bar{f})^2}\sqrt{\sum_{t=1}^{n}(f(s_{t+k})-\bar{f})^2}},
\end{equation*}

\noindent
where $\bar{f} = \frac{1}{n}\sum_{t=1}^n f(s_{t})$. 

\

Given the huge size of the neuroevolution search space, and in the attempt to generate walks that are, as much as possible, representatives of the
portions of the search space actually explored by the evolutionary algorithm,
in this work we have decided to calculate autocorrelation using {\it selective walks}~\cite{thesis}.
In selective walks, for each $t = 1,2,...,n$, $s_{t}$ is a selected solution from the neighborhood of~$s_{t-1}$.
To apply selection pressure to the neighbors, tournament selection is used;
in other words, $s_{t}$ is the best solution (i.e.,~the one with the best fitness on the training set)
in a sample of~$m$  randomly generated neighbors of $s_{t-1}$.
We study the autocorrelation both on the training and on the test set,
by using the same selective walk.
In both cases selection acts using only training data, but in the former case the individuals are evaluated on the training set, while in the latter case they are evaluated on the test set.

Because of the large complexity of neuroevolution,
and given the relatively short length of the walks that we are able to
generate with the available computational resources\footnote{Our experiments were performed on a gtx~970 and on a gtx~2070.}~($n=30$~in our experiments),
we have decided to calculate~$\hat{\rho}(k)$ several times~(10~in our experiments),
using independent selective walks, and to report boxplots of the results
obtained over these different walks.
In order to broadly classify the ruggedness of the landscape, we have adopted 
the heuristic threshold suggested by Jones for fitness-distance correlation~\cite{jones95a}: 
$\hat{\rho}(k) > 0.15$ will correspond to a smooth landscape (and thus, in principle,
an easy problem), while $\hat{\rho}(k) < 0.15$ will correspond to a 
hard landscape.
To visualize the results, the threshold will be shown as an horizontal line
in the same diagram as the boxplots of the autocorrelation, and the
position of the box compared to the threshold will allow us to classify
problems as easy or hard.
The situation in which the boxplot lays across the threshold
(i.e., the case in which~$\hat{\rho}(k) \approx 0.15$) 
will be considered as an {\it uncertain} case, in which
predicting the hardness of the problem is difficult.
One of the typical situations in which we have an uncertain case is when several different neuroevolution runs give significantly different outcomes (for instance, half of the runs converge towards good quality solutions and the other half stagnate in bad quality ones).
Finally, several values of the step~$k$ are compared~($k = 1,2,3,4$ in our
experiments).

%
%
% The correlation length $\tau$ measures how the autocorrelation function decreases, with 
% larger values representing smoother surfaces. According to the definition proposed by 
% Weinberger~\cite{Weinberger1990}, $\tau = - \frac{1}{\ln(\rho(1))}$
% $<$IF WE DON'T USE IT, THERE IS NO REASON TO DEFINE IT. 
% LET'S THINK ABOUT IT$>$.
%
%

%%%%%%%%%%%%%%%%%%%%%%%%%%%%%
%
%   Entropic Measure of Ruggedness
%
%%%%%%%%%%%%%%%%%%%%%%%%%%%%%
\subsection{Entropic Measure of Ruggedness}
\label{entrmeasofrug}

The EMR is 
an indicator of the relationship between ruggedness and neutrality. 
It was introduced by Vassilev~\cite{Vassilev:phd,Vassilev2000,Vassilev2003} and is defined as:
assuming that a walk of length $n$, performed on a landscape, generates a time series of fitness values $\{f_{t}\}^n_{t=0}$, that time series can be represented as a string $S(\varepsilon) = \{x_{1}x_{2}...x_{n}\}$, where, for each~$i=1,2,...,n$, $x_{i} \in \{ \bar{1}, 0, 1\}$. 
For each~$i=1,2,...,n$, $x_i \in S(\varepsilon)$ is obtained using the following function:

\begin{equation*}
    x_{i} = \Psi_{f_{t}}(i, \varepsilon) = 
    \begin{cases}
        \bar{1}, \; \mbox{if } f_{i} - f_{i-1} < -\varepsilon          \\
        0,       \; \mbox{if } |f_{i} - f_{i-1}| \leq -\varepsilon                   \\
        1,       \; \mbox{if } f_{i} - f_{i-1} > -\varepsilon
    \end{cases}
\end{equation*}

%\begin{figure}[H]
%\centering
%    \includegraphics[scale=.45]{pic/entropy.png}
%    \caption{Classification and encoding of three-point objects}
%\end{figure}

\noindent
where
$\varepsilon$ is a real number that determines the accuracy of the calculation of $S(\varepsilon)$, and increasing this value results in increasing the neutrality of the landscape. The smallest possible~$\varepsilon$ for which the landscape becomes flat is called the information stability, and is represented by $\varepsilon^*$.
Using $S(\varepsilon)$, the EMR is defined as follows\cite{Vassilev:phd}:

\begin{equation*}
    H(\varepsilon) = - \sum_{\substack{p \neq q}} P_{[pq]}\log_6 P_{[pq]},
\end{equation*}

\noindent
where $p$ and $q$ are elements from the set $\{\bar{1}, 0, 1\}$, and 
\mbox{$P_{[pq]} = \frac{n_{[pq]}}{n}$},
where $n_{[pq]}$ is the number of $pq$~sub-blocks in~$S(\varepsilon)$ and~$n$ is the total number of sub-blocks.
The output of $H(\varepsilon)$ is a value in the $[0,1]$ range, and it represents an estimate of the variety of fitness values in the walk, with a higher value meaning a larger variety and thus a more rugged landscape.
In this definition, $H(\varepsilon)$ is calculated for multiple $\varepsilon$ values, usually %$\{0, \frac{\varepsilon^*}{128}, \frac{\varepsilon^*}{64}, \frac{\varepsilon^*}{32}, \frac{\varepsilon^*}{16}, \frac{\varepsilon^*}{8}, \frac{\varepsilon^*}{4}, \frac{\varepsilon^*}{2}, \varepsilon^* \}$,
$\{0, \varepsilon^*/128, \varepsilon^*/64, \varepsilon^*/32, \varepsilon^*/16, \varepsilon^*/8, \varepsilon^*/4, \varepsilon^*/2, \varepsilon^* \}$,
then a mean $\bar{H}(\varepsilon)$ of each $H(\varepsilon)$ is calculated and used as the value to be analysed.
In this work, we employ the adaptations suggested by Malan~\cite{Malan}, aimed at reducing the characterization of the landscape to a single scalar.
To characterise the ruggedness of a function~$f$, the following value is proposed:

\begin{equation*}
    R_{f} = \max_{\forall \varepsilon \in [0, \varepsilon^*]} {H(\varepsilon)}
\end{equation*}

\noindent
To approximate the theoretical value of $R_{f}$, the maximum of $\bar{H}(\varepsilon)$ is calculated for all $\varepsilon$ values.
%
%
%
% Another important modification proposed by the same authors was regarding the definition of $S(\varepsilon)$. In the original definition~\cite{Vassilev2000}, there is the assumption that the landscape is statistically isotropic, so a walk of $n$ steps results in a $S(\varepsilon)$ of $n$ symbols because the last fitness value is compared back to the first value. For continuous landscapes, this definition was modified to ``not wrap around'', and so a walk of $n$ steps results in a $S(\varepsilon)$ of $n-1$ symbols.

%%%%%%%%%%%%%%%%%%%%%%%%%%%%%
%
%   Neuroevolution
%
%%%%%%%%%%%%%%%%%%%%%%%%%%%%%
\section{Neuroevolution}
\label{ne}

Neuroevolution is usually employed to evolve the topology, weights, parameters and/or learning strategies of artificial neural networks.
Some of the most well known neuroevolution systems include EPNet~\cite{epnet}, NEAT~\cite{neat}, EANT~\cite{eant}, and hyperNEAT~\cite{hyperneat}.
% EPNet presented a novel way to evolve network behaviours using evolutionary programming, NEAT demonstrated how advantageous it was to evolve both weights and the network architecture, EANT proposed a new method to both exploit and explore network structures, and hyperNEAT demonstrated how indirect encoding strategies could be used to describe connectivity patterns in very large networks.
Most recently, works have appeared that apply neuroevolution to other types of neural networks, such as 
%convolutional neural networks
CNNs~\cite{hneatcnn,dppn,codeapneat}. 
% Such works include the proposed modification to hyperNEAT, that allowed for the evolution of CNNs~\cite{hneatcnn}, a new version of CPPNs, called DPPNs~\cite{dppn}, that are able to replicate CNN architectures, and  CoDeapNEAT~\cite{codeapneat}, which was able to evolve convolutional and recurrent networks.
%
%
%
%
%
%
In this section we describe how we represent networks 
using a grammar-based approach, and we discuss 
the employed genetic operations.

%---------------------------------
%
%   Grammar-based neuroevolution
%
%---------------------------------
\subsection{Grammar-Based Neuroevolution}
\label{subsec:gbn}

We have decided to use a grammar-based approach because of its modularity and flexibility.
The employed grammar is reported in Fig.~\ref{fig:grammar}.
It contains all the possible values for the parameters of each available layer. 
This way, adding and removing layers or changing their parameters
is simple and requires minimal changes. 
%--------------------------------
\begin{figure}[!ht]
\centering
  \begin{small}   
    \begin{tabular}{lll}
        Conv ::      & filters        & \textbar 32,64,128,256\textbar                \\
                     & kernel\_size   & \textbar 2,3,4,5\textbar                      \\
                     & stride         & \textbar 1,2,3\textbar                        \\
                     & activation     & \textbar relu, elu, sigmoid\textbar           \\
                     & use\_bias      & \textbar true, false\textbar                  \\
        Pool ::      & type           & \textbar Max, Avg\textbar                     \\
                     & pool\_size     & \textbar 2,3,4,5\textbar                      \\
                     & stride         & \textbar 1,2,3\textbar                        \\
        Dense ::     & units          & \textbar 8,16,32,64,128,256,512\textbar       \\
                     & activation     & \textbar relu, elu, sigmoid\textbar           \\
                     & use\_bias      & \textbar true, false\textbar                  \\
        Dropout ::   & rate           & [0.0 $\to$ 0.7]                       \\
        Optimizer :: & learning\_rate & \textbar 0.01, 0.001, 0.0001, 0.00001\textbar \\
                     & decay          & \textbar 0.01, 0.001, 0.0001, 0.00001\textbar \\
                     & momentum       & \textbar 0.99, 0.9, 0.5, 0.1\textbar          \\
                     & nesterov       & \textbar true, false\textbar                 
    \end{tabular}
   \end{small}
    \caption{Grammar used to evolve the CNNs.}
    \label{fig:grammar}
\end{figure}
%------------------------------------
Using this grammar, we are discretizing the range of the possible values that each parameter can take. This greatly reduces the search space, while keeping the quality of the solutions under control, as in most cases, intermediate values can have little to no significant influence on the effectiveness of the solutions, as reported in~\cite{BALDOMINOS}.
In our representation,
genotypes are composed by two different sub-sections, $S_1$ and $S_2$, that are connected using the so called Flatten gene. The Flatten gene implements the conversion~(i.e.,~the ``flattening'') of multidimensional arrays into unidimensional arrays, an operation required to connect the convolutional 
and the fully connected layers.
On~$S_1$ we have genes that encode the layers that deal with feature extraction from the images, convolutional and pooling layers, and on~$S_2$ we have genes that encode classification and pruning, dense and dropout layers.
This separation allows for a more balanced generation of the genomes and application of the genetic operators.
Besides the flatten layer, the only other layer that is the same for all possible individuals is the output layer, which is a fully connected (i.e., dense) layer
 with softmax activation and a number of units equal to the number of classes to be predicted.
The genetic operators cannot modify this layer, except for the bias parameter.

Before evaluation, a genotype is mapped into a phenotype, that is a neural network itself.
Evaluation involves training the network and calculating its performance on the given data.
During the evolutionary process, we use the loss value on the training set as a fitness function to evaluate the networks.
Regarding the optimizer used for training the networks, we have chosen Stochastic Gradient Descent~(SGD)~\cite{wilson2017marginal}.
%
%
%    due to two important factors:
%   \begin{itemize}
%       \item The range and possible values for the SGD parameters have been extensively 
%       tested, so the value choice could be done more accurately.
%      \item Although ADAM~\cite{adam} is more common nowadays due to achieving 
%       better overall results, SGD outperforms ADAM when it comes to generalization 
%       ability~\cite{wilson2017marginal}.
%   \end{itemize}
%
%
Also, since we are working with multiclass classification problems that are not one-hot encoded, we used Sparse Categorical Cross-Entropy as a loss function, which motivates the need to have the fixed number of neurons and activation function in the output layer.
%
%  Since the objective of the evolutionary process is to confirm the results 
%  from the selective walk, we do not apply any form of elitism between generations.
%  Each generation $G_n$ will consist of the off-springs generated 
%  by generation $G_{n-1}$.
%  The parents are selected for reproduction using tournament selection.
%  No elitism is used.
%
%
We also measure the accuracy and loss in a separate 
test set in order to study the generalization ability.

%----------------------------------------
% \begin{figure}[!ht]
% \centering
%    \includegraphics[width=0.4\textwidth]{pics/network.png}
%    \caption{Genotype example}
%    \label{fig:genotype}
% \end{figure}
%----------------------------------------

%--------------------------------------
%
%  Genetic Operators
%
%--------------------------------------
% \subsection{Genetic Operators}
% \label{genop}
%

Due to the difficulty 
of defining a crossover-based neighborhood for studying FLs~\cite{crossoverdistance},
we consider only mutation.
Having in mind the vast amount of possible mutation choices, we have decided to 
restrict our study to three different types of operators:

\begin{itemize}
    \item Topology mutations. Mutations that add or delete a gene, except for the flatten gene, changing the topology of the genotype and, consequently, the one of the phenotype.

    \item Parameter mutations. Mutations that change the parameters encoded in a gene. They cover all parameters of all gene types, excluding the flatten gene.

    \item Learning mutations. Mutations that change the parameters related to the learning process, which are encoded in the Optimizer gene.
\end{itemize}

%%%%%%%%%%%%%%%%%%%%%%%%%%%%%
%
%   Experimental Study
%
%%%%%%%%%%%%%%%%%%%%%%%%%%%%%
\section{Experimental Study}
\label{estud}

%%%%%%%%%%%%%%%%%%%%%%%%%%%%%
%
%   Datasets and  Experimental Settings
%
%%%%%%%%%%%%%%%%%%%%%%%%%%%%%
\subsection{Datasets and  Experimental Settings}
\label{deset}

%
%   We selected 4 different multiclass classification problems, 
%   with different degrees of difficulty.
% 

Table~\ref{fig:dsets} describes the main characteristics of the datasets used as test cases in our experiments. The partition into training and test set is made randomly, and it is different at each run.
For all datasets, a simple image scale adjustment was done, setting pixel values in the $[0,1]$ range. No further data pre-processing or image augmentation was applied to the datasets.
The MNIST dataset consists in a set of gray scale images of handwritten digits from~0 to~9~\cite{mnist}.
Fashion-MNIST (FMNIST) is similar to MNIST, but instead of having digits from~0 to~9, it contains images of~10 different types of clothing articles~\cite{fashionmnist}.
CIFAR10 contains RGB pictures of 10 different types of real world objects~\cite{cifar10}.
Finally, SVHN contains RGB pictures of house numbers, containing digits from~0 to~9~\cite{svhn}.
For each one of these four datasets
and for each one of the three studied mutation operators, 
we perform selective walks (that allow us to have all the needed information to calculate the autocorrelation and the~EMR) and we execute the neuroevolution.
From now on, we will use the term {\it configuration} to indicate an experiment in which a particular type of mutation
was used on a particular dataset. 
For each configuration, we perform 10 independent selective walks and 10 independent neuroevolution runs.
%% Results reported for the autocorrelation 
%% are boxplots of the value of the measure
%% obtained on this 10 independent runs.
All neuroevolution runs are performed starting with a randomly initialized population of individuals, and all the selective walks are constructed starting with a randomly generated individual.

To determine the values of the main parameters (e.g., population size and number of generations for neuroevolution, length of the walk and number of neighbors for selective walks) we have performed some benchmark tests with multiple values, and selected ones that allowed us to obtain results in ``reasonable'' time\footnote{On average, 5 hours per run.} with our available computational resources. The employed parameter values 
are reported in Table~\ref{fig:evoparams}.
The first two columns contain the parameters used to build the selective walks and the parameters of the neuroevolution, respectively. One should keep in mind that, in order to evaluate all the neural networks in the population, all the networks need to go through a learning phase at each generation of the evolutionary process. So, the third column reports the values used by each one of those networks for learning. When decoding the genotype into the phenotype, the weights of the network are randomly initialized using the Xavier initialization~\cite{Glorot10understandingthe}.

%------------------------------------------------
%  2 tables together
%------------------------------------------------
\begin{table}[t]
    \centering
    
    \caption{Number of training and test observations, and number of classes of each dataset.}
    \label{fig:dsets}
\begin{tabular}{c|ccc}
 & \textbf{Training set} & \textbf{Test set} & \textbf{Classes} \\ \hline
MNIST   & 50000 & 10000 & 10 \\
FMNIST  & 50000 & 10000 & 10 \\
CIFAR10 & 50000 & 10000 & 10 \\
SVHN    & 73257 & 26032 & 10       
\end{tabular}

% WHY THIS WAS MULTICOLUMN-COMPLEX TABLE?
%\begin{tabular}{lccc}
                            %       & %\multicolumn{1}{l}{Training %set} & \multicolumn{1}{l}{Test %set} & %\multicolumn{1}{l}{Classes} \\ %\hline
%\multicolumn{1}{l|}{MNIST}         & 50000                    %        & 10000                        & 10                   %       \\
%\multicolumn{1}{l|}{FMNIST} & 50000                           % & 10000                        & 10                          %\\
%\multicolumn{1}{l|}{CIFAR10}       & 50000                    %        & 10000                        & 10                   %       \\
%\multicolumn{1}{l|}{SVHN}          & 73257                    %        & 26032                        & 10       
%\end{tabular}

\bigskip
\bigskip

    \caption{Parameter values used in our experiments.}
    % The first column contains the parameters used to generate the selective walks. The second column shows the parameters used by neuroevolution. The third column reports the parameters used for learning by each neural network in the population.
    \label{fig:evoparams}
\begin{tabular}{|l@{~}r|l@{~}r|lr|}
%\hline
\multicolumn{2}{|c|}{\textbf{~~Selective walk~~}}  & \multicolumn{2}{c|}{\textbf{Neuroevolution}}  & \multicolumn{2}{c|}{\textbf{Learning}} \\ \hline \noalign{\smallskip}
~Length       & 30~ & Population size & 10 & ~~~Epochs & 8~~          \\
~\# Neighbors & 3~  & \# Generations  & 20 & ~~~Batch  & 64~~        \\
             &    & Tournament size & 2  &        &           \\
             &    & Mutation rate   & 1  &        &           \\
             &    & Crossover rate  & 0  &        &           \\
             &    & No elitism      &    &        &           \\
%                     \hline 
\end{tabular}

% ESTAVA FEIA!!!
%\begin{tabular}{|lr|l|lr|l|lr|}
%\hline
%\multicolumn{2}{|c|}{\textbf{Selective walk}} &  & \multicolumn{2}{c|}{\textbf{Neuroevolution}} &  & \multicolumn{2}{c|}{\textbf{Learning}} \\ \hline
%Length               & 30           &  & Population size       & 10        &  & Epochs          & 8           \\
%\# Neighbors         & 3            &  & \# Generations        & 20        &  & Batch           & 64          \\
%                     &              &  & Tournament size         & 2         &  &                 &           \\
%                     &              &  & Mutation rate        & 1         &  &                 &           \\
%                     &              &  & Crossover rate         & 0         &  &                 &           \\
                     %&              &  & No elitism         & %        &  &                 &           %\\
%                     \hline 
%\end{tabular}

\end{table}

%-----------------------------------------------------------------------------------------
%  \begin{table}[ht]
%  \centering
%      \caption{Evolutionary parameters}
%      \label{fig:evoparams}
%  \begin{tabular}{lll}
%  \hline
%  \multicolumn{3}{c}{Neuroevolution}      \\ \hline
%  Population       &      & 10    \\
%  Generations      &      & 20    \\ 
%  Tournament size  &      & 2    \\ \hline
%  \multicolumn{3}{c}{Selective Walk} \\ \hline
%  Length           &      & 30       \\
%  Neighbors        &      & 3        \\ \hline
%  \multicolumn{3}{c}{Learning} \\ \hline
%  Epochs           &      & 8       \\
%  Batch size       &      & 64        \\
%  \end{tabular}
%  \end{table}
%
%
%----------------------------------------------

%%%%%%%%%%%%%%%%%%%%%%%%%%%%%
%
%   Experimental Results
%
%%%%%%%%%%%%%%%%%%%%%%%%%%%%%
\subsection{Experimental Results}
\label{er}

\begin{figure*}[!h]
    \captionsetup[subfigure]{justification=centering}
        \centering
        \begin{subfigure}{0.31\textwidth}
        \hspace*{0.2cm}
            \includegraphics[width=\textwidth]{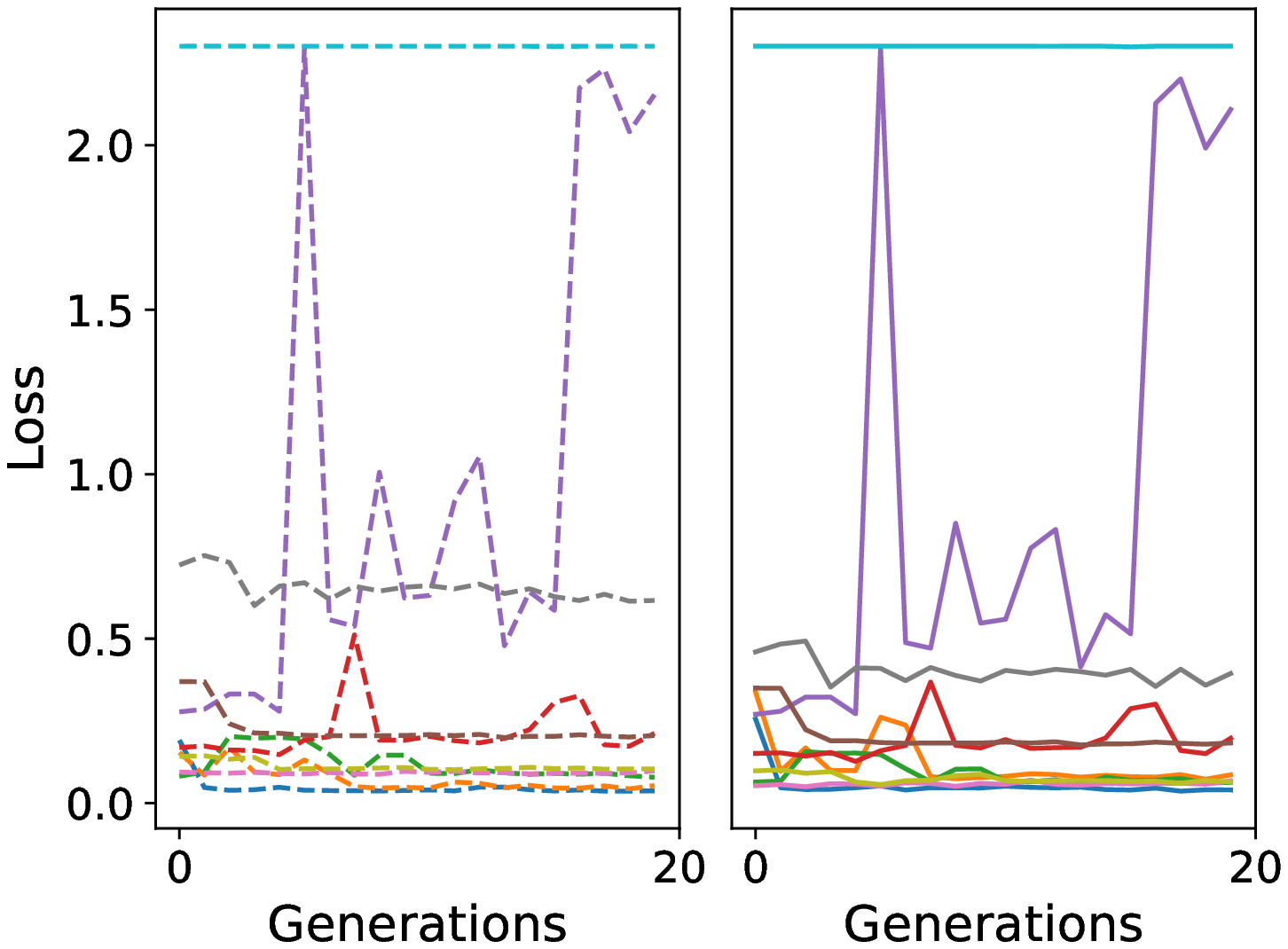}
                \caption{Learning}
                \label{fig:evo-mnist-learning}
            \end{subfigure}
            \begin{subfigure}{0.31\textwidth}
            \hspace*{0.2cm}
            \includegraphics[width=\textwidth]{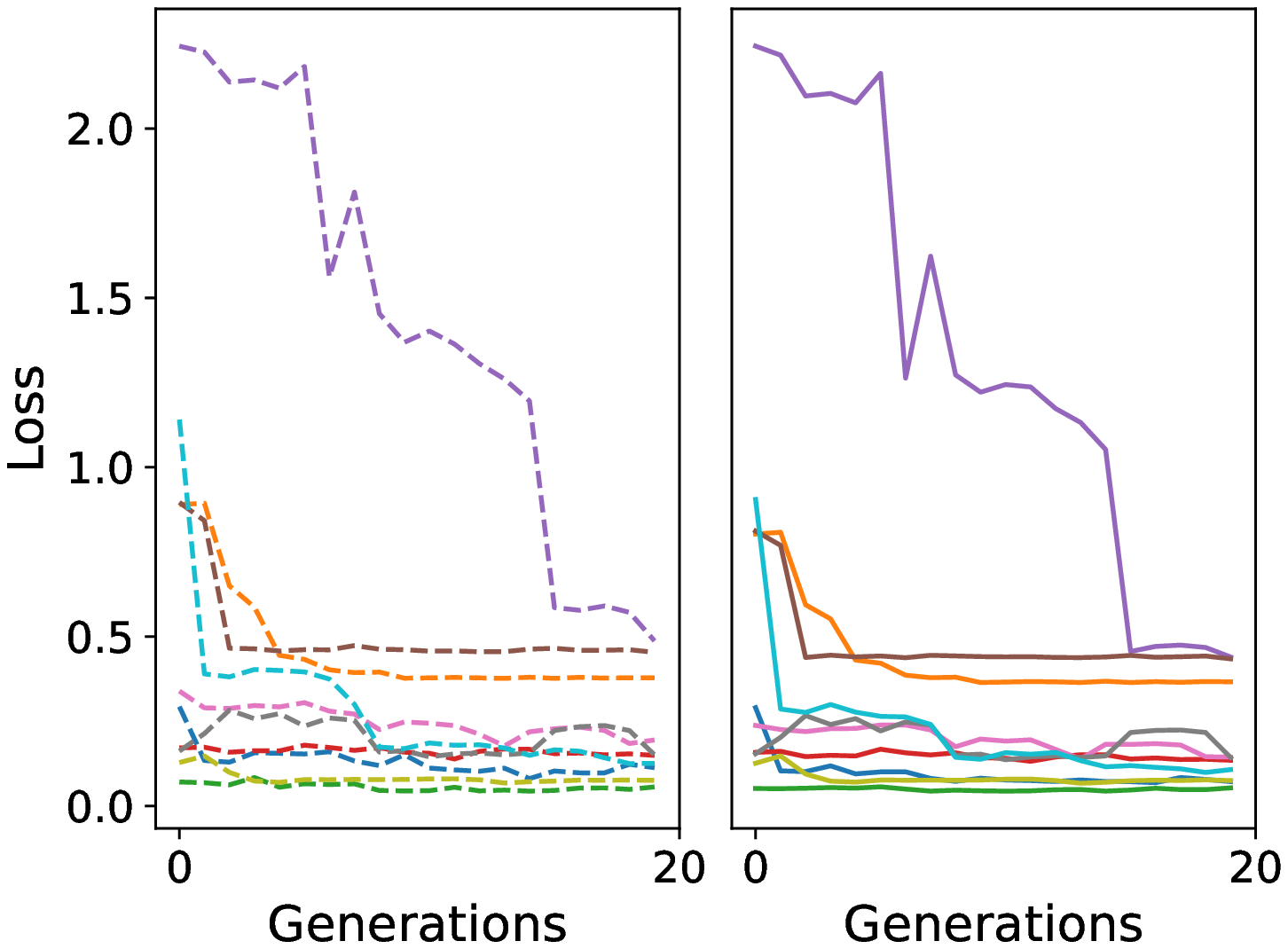}
                \caption{Parameters}
                \label{fig:evo-mnist-params}
            \end{subfigure}
            \begin{subfigure}{0.31\textwidth}
            \hspace*{0.2cm}
            \includegraphics[width=\textwidth]{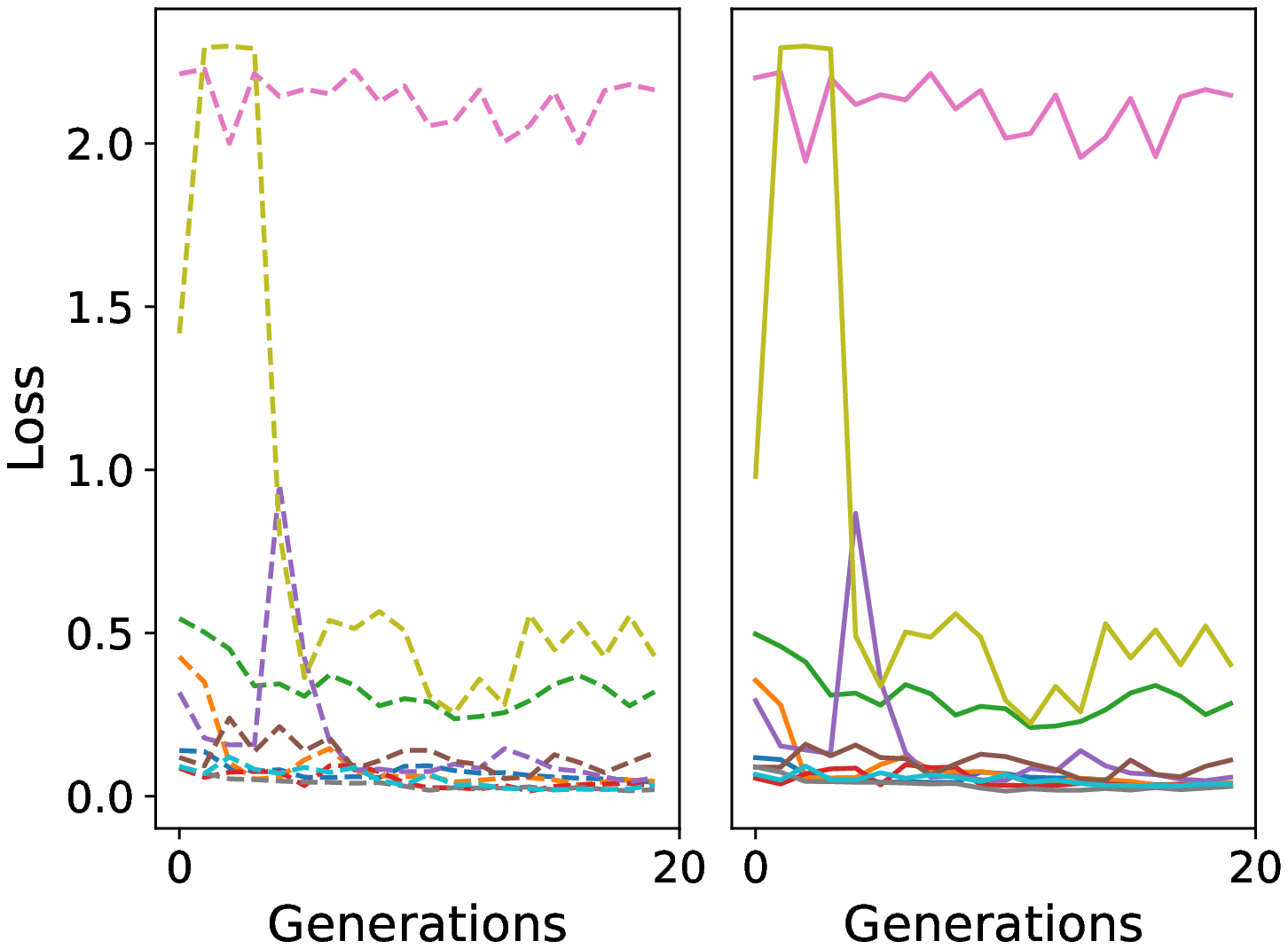}
                \caption{Topology}
                \label{fig:evo-mnist-topology}
            \end{subfigure}

          \begin{subfigure}{0.31\textwidth}
            \includegraphics[width=\textwidth]{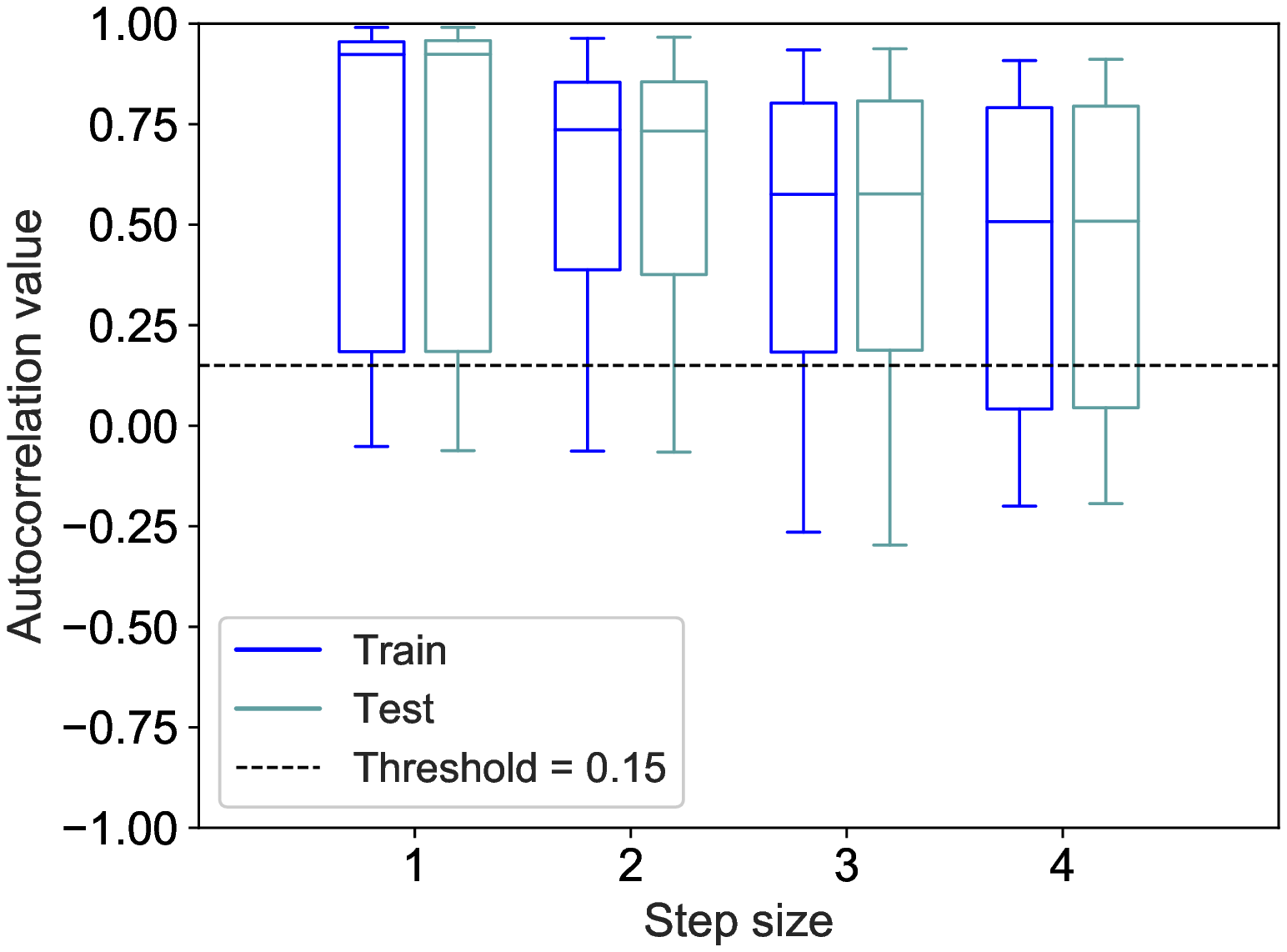}
              \caption{Learning}
              \label{fig:ac-mnist-learning}
          \end{subfigure}
          \begin{subfigure}{0.31\textwidth}
            \includegraphics[width=\textwidth]{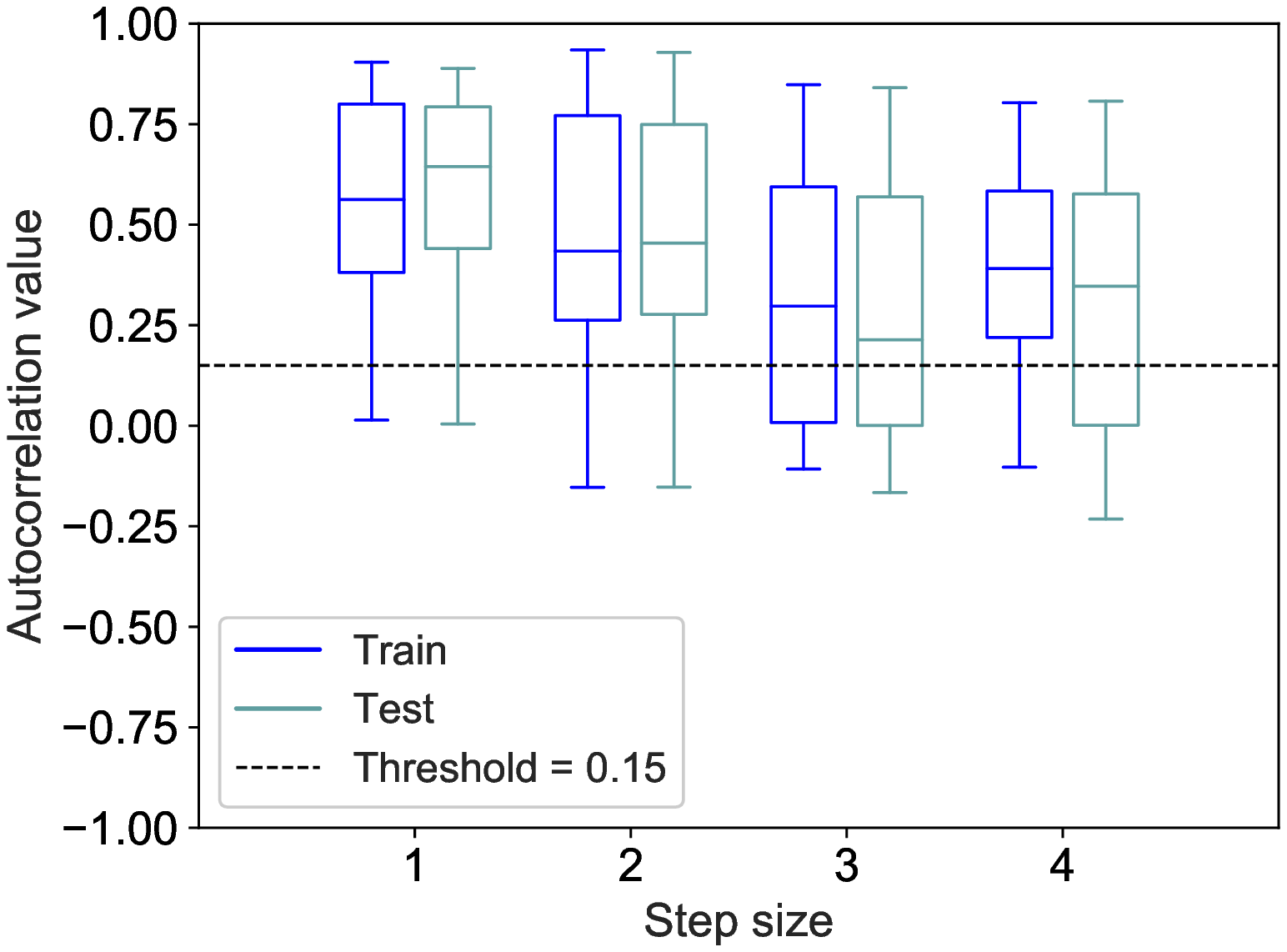}
              \caption{Parameters}
              \label{fig:ac-mnist-params}
          \end{subfigure}
          \begin{subfigure}{0.31\textwidth}
            \includegraphics[width=\textwidth]{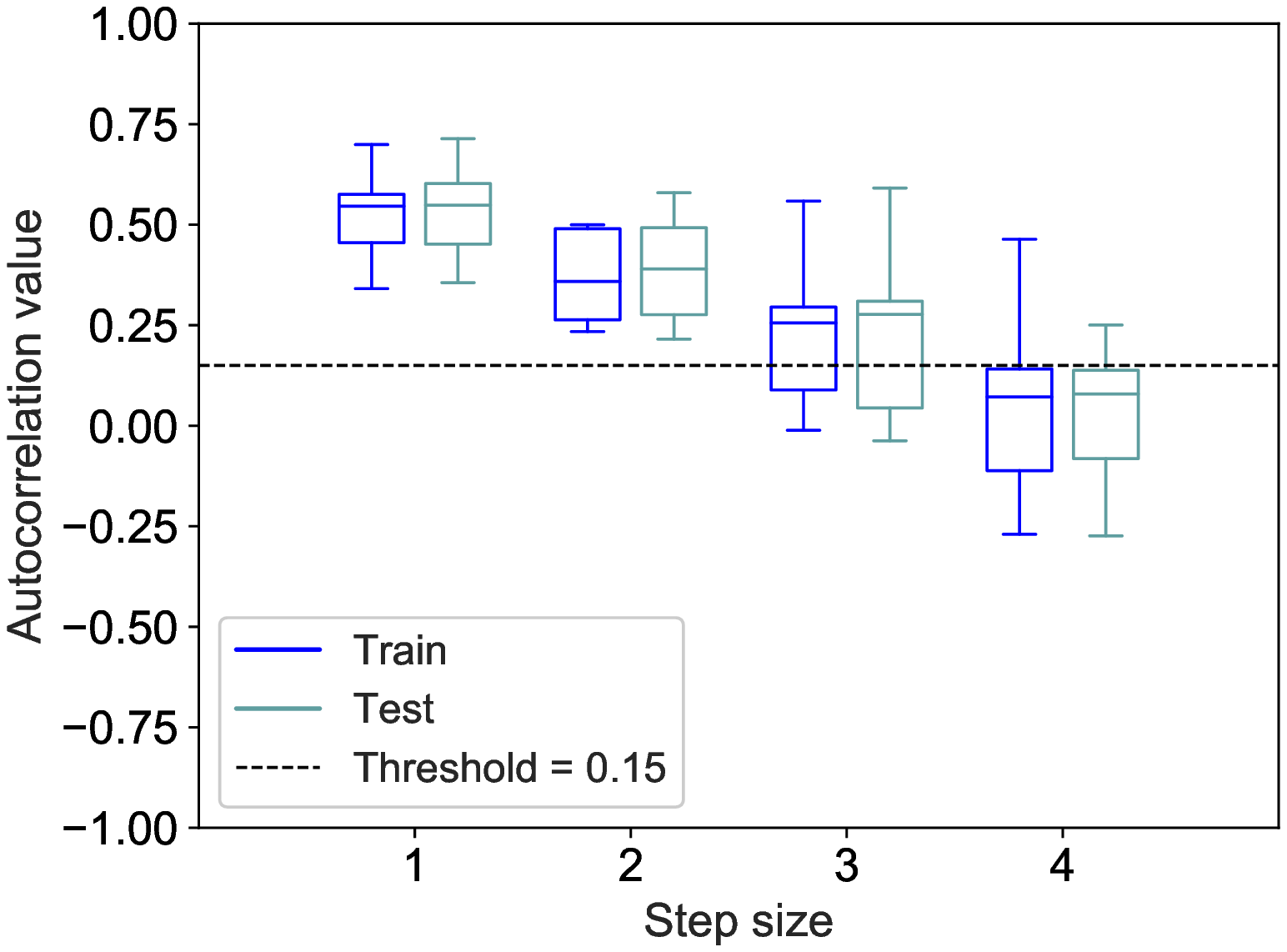}
              \caption{Topology}
              \label{fig:ac-mnist-topology}
          \end{subfigure}
    \caption{{\it MNIST dataset}. Plots~(a), (b) and~(c):~neuroevolution results; 
    plots~(d), (e) and~(f):~autocorrelation results.
    Plots~(a), (b) and~(c) report the evolution of the best fitness in the population
    at each generation (one curve for each performed neuroevolution run). 
    Each plot is partitioned into two subfigures: training loss on the left and 
    test loss on the right.
    Plots~(d), (e) and~(f) report the boxplots of the autocorrelation, calculated
    over 10 independent selective walks.}
    \label{fig:ac-mnist}
\end{figure*}

We begin by analyzing the ability of autocorrelation to characterize training and test performance of neuroevolution of~CNNs.
Fig.~\ref{fig:ac-mnist} reports the evolution of the loss and the autocorrelation for the MNIST problem.
%--------------------------------------------------------------------------------------------
%   
%     BEGIN FIGURE NEUROEVOLUTION VS AUTOCORRELATION
%                                      MNIST
%
%--------------------------------------------------------------------------------------------
%--------------------------------------------------------------------------------------------
%   
%     END FIGURE NEUROEVOLUTION VS AUTOCORRELATION
%                                      MNIST
%
%--------------------------------------------------------------------------------------------
The first line of plots reports the evolution of the loss 
against generations for the three studied mutation operators. 
Each plot in the first line is partitioned into two halves: the leftmost one
reports the evolution of the training loss of the best individual, while the rightmost one reports the loss of the same individual,
evaluated on the unseen test set.
Each curve in these plots reports the results of one neuroevolution run.
The second line contains the boxplots of the autocorrelation values,
calculated over 10 independent selective walks, both on the training
and on the test set.
Each column of plots reports the results 
for a different type of mutation, allowing us to easily compare the outcome
of the neuroevolution and the one of the autocorrelation for the different configurations.

As we can observe from plots~(a) and~(b) of Fig.~\ref{fig:ac-mnist}, 
when we employ learning mutation
and parameters mutation, the MNIST problem is easy to solve, both on training
and test set.
In fact, except for the outlier runs in plot~(a), all the runs either approximate 
the ideal value of loss equal to zero, or tend towards it.
Now, looking at plots~(d) and~(e), it is possible to observe that the autocorrelation
is able to capture the fact that the problem is easy.
In fact, in both cases, practically the whole autocorrelation box stands above (and the medians never go below) the~0.15 threshold.
When the topology mutation is used, the situation changes: the number of runs in which the evolution does not have a regular trend is larger.
%, and there are cases in which neuroevolution is not able to get close to a loss equal to zero.
This may not be obvious by looking at plot~(c) because of the scale of the y axis, but the lines are now much more rugged than they were for the other two cases. The problem is now harder than it was, and as we can see in plot~(f), the autocorrelation catches 
this difficulty. In particular, we can observe that when the step is equal to~4,
the whole autocorrelation boxes are below the threshold.
%, and also when the step is equal to~3, the lower parts of the boxes stand across the threshold.
% THIS ALSO HAPPENS FOR Parameters, so let's no say it here.
%
%
%
The partial conclusion that we can draw for the MNIST dataset is that learning and parameter mutations
are more effective operators than topology mutations, and this is correctly predicted
by the autocorrelation.
Furthermore, we can observe that the neuroevolution results obtained on the test set are
very similar to the ones on the training set, practically for all the runs we have
performed. Also this feature is captured by the autocorrelation, since the training and test boxes are very similar
to each other for practically all the configurations.

\begin{figure*}[!h]
    \captionsetup[subfigure]{justification=centering}
        \centering
        \begin{subfigure}{0.31\textwidth}
        \hspace*{0.2cm}
            \includegraphics[width=\textwidth]{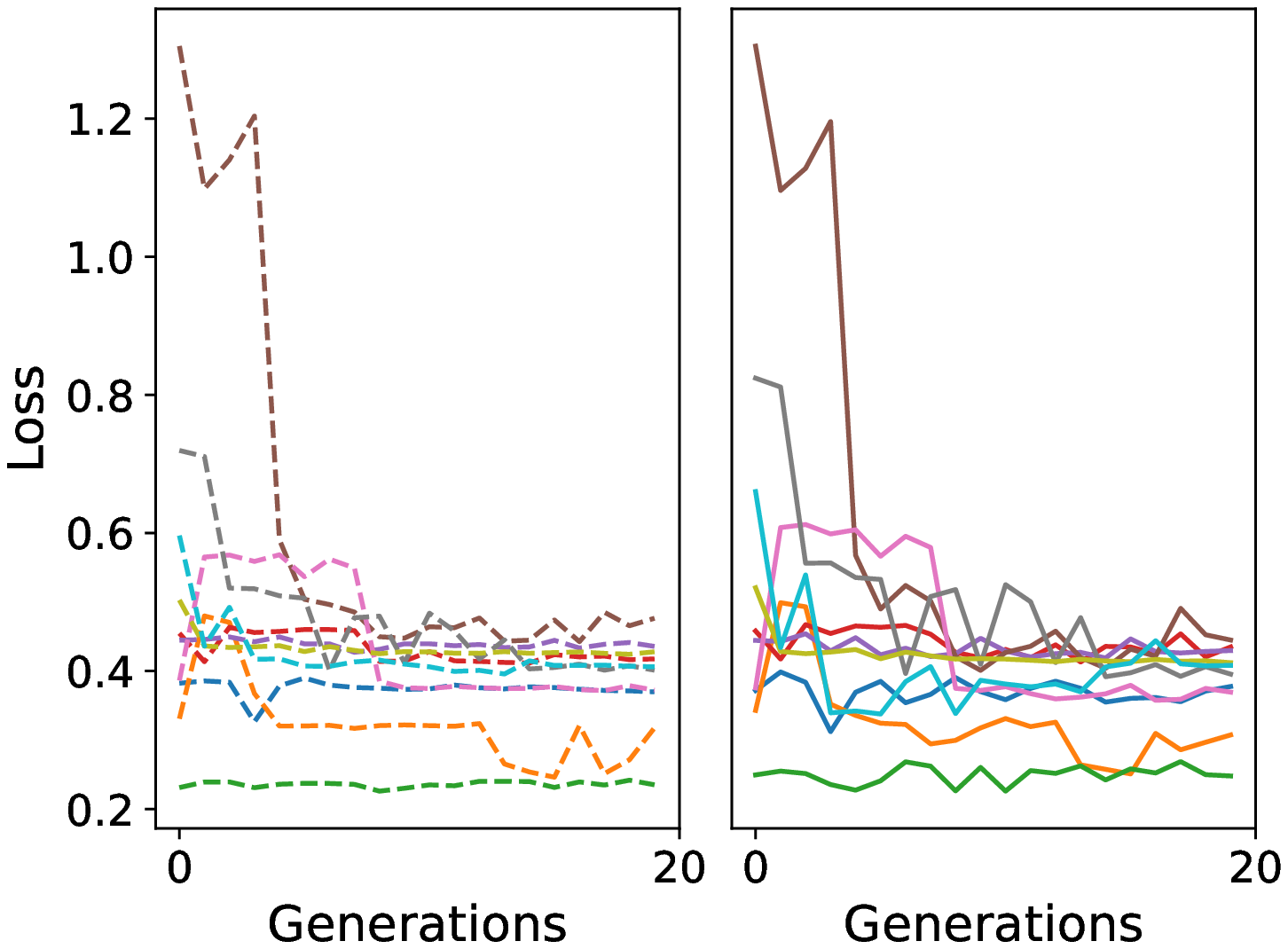}
                \caption{Learning}
                \label{fig:evo-fmnist-learning}
            \end{subfigure}
            \begin{subfigure}{0.31\textwidth}
            \hspace*{0.2cm}
            \includegraphics[width=\textwidth]{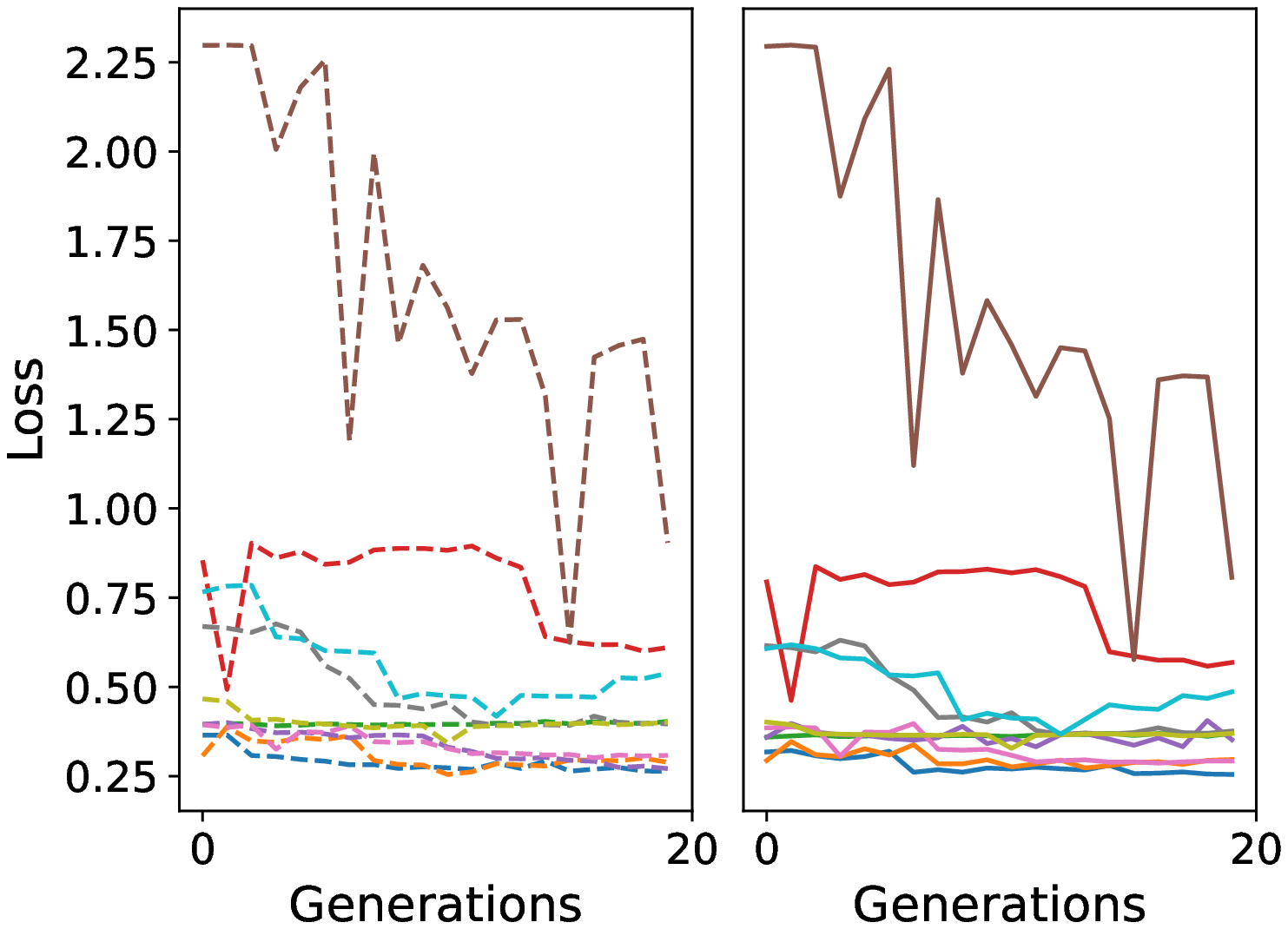}
                \caption{Parameters}
                \label{fig:evo-fmnist-params}
            \end{subfigure}
            \begin{subfigure}{0.31\textwidth}
            \hspace*{0.2cm}
            \includegraphics[width=\textwidth]{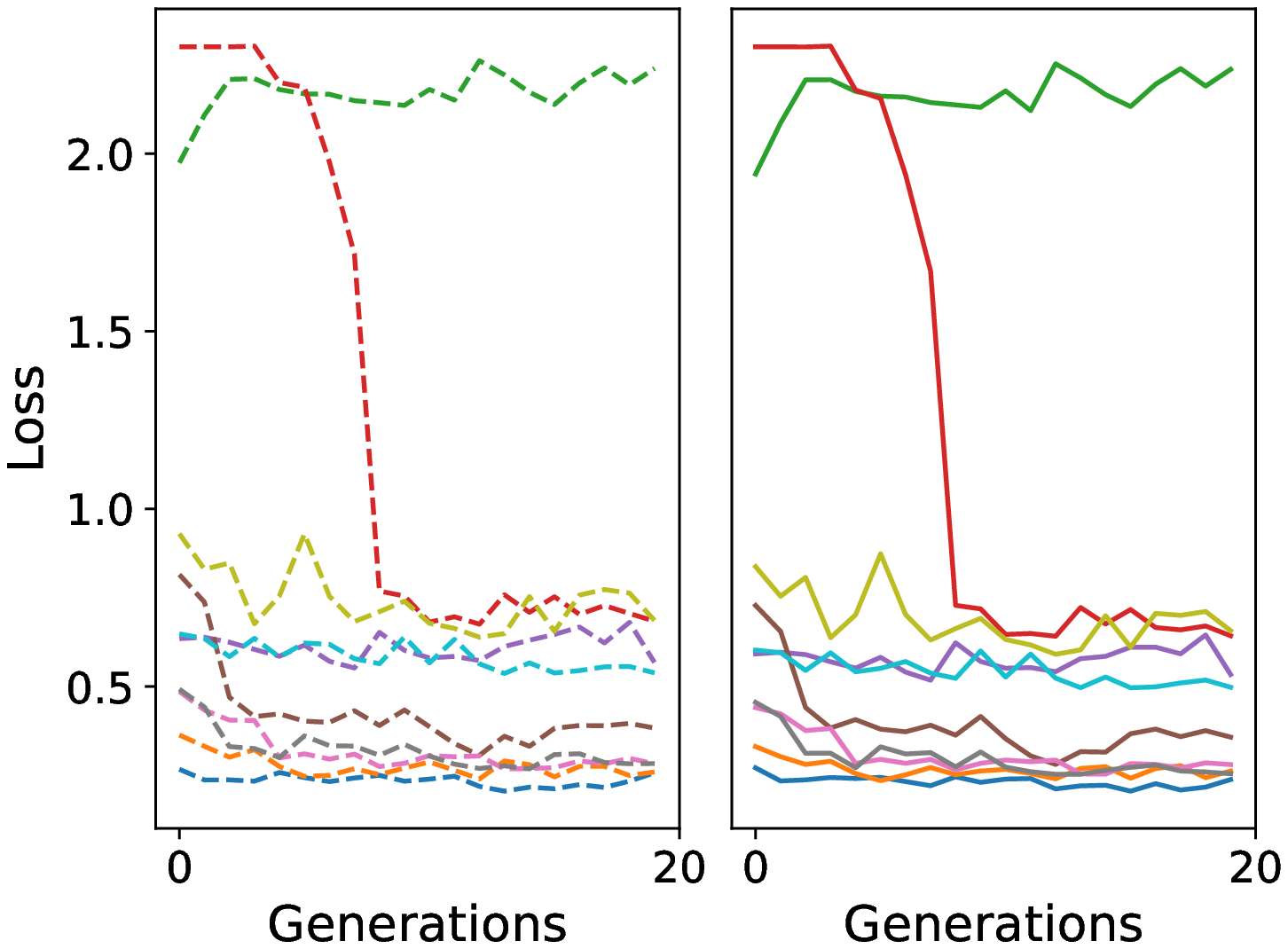}
                \caption{Topology}
                \label{fig:evo-fmnist-topology}
            \end{subfigure}

            \begin{subfigure}{0.31\textwidth}
            \includegraphics[width=\textwidth]{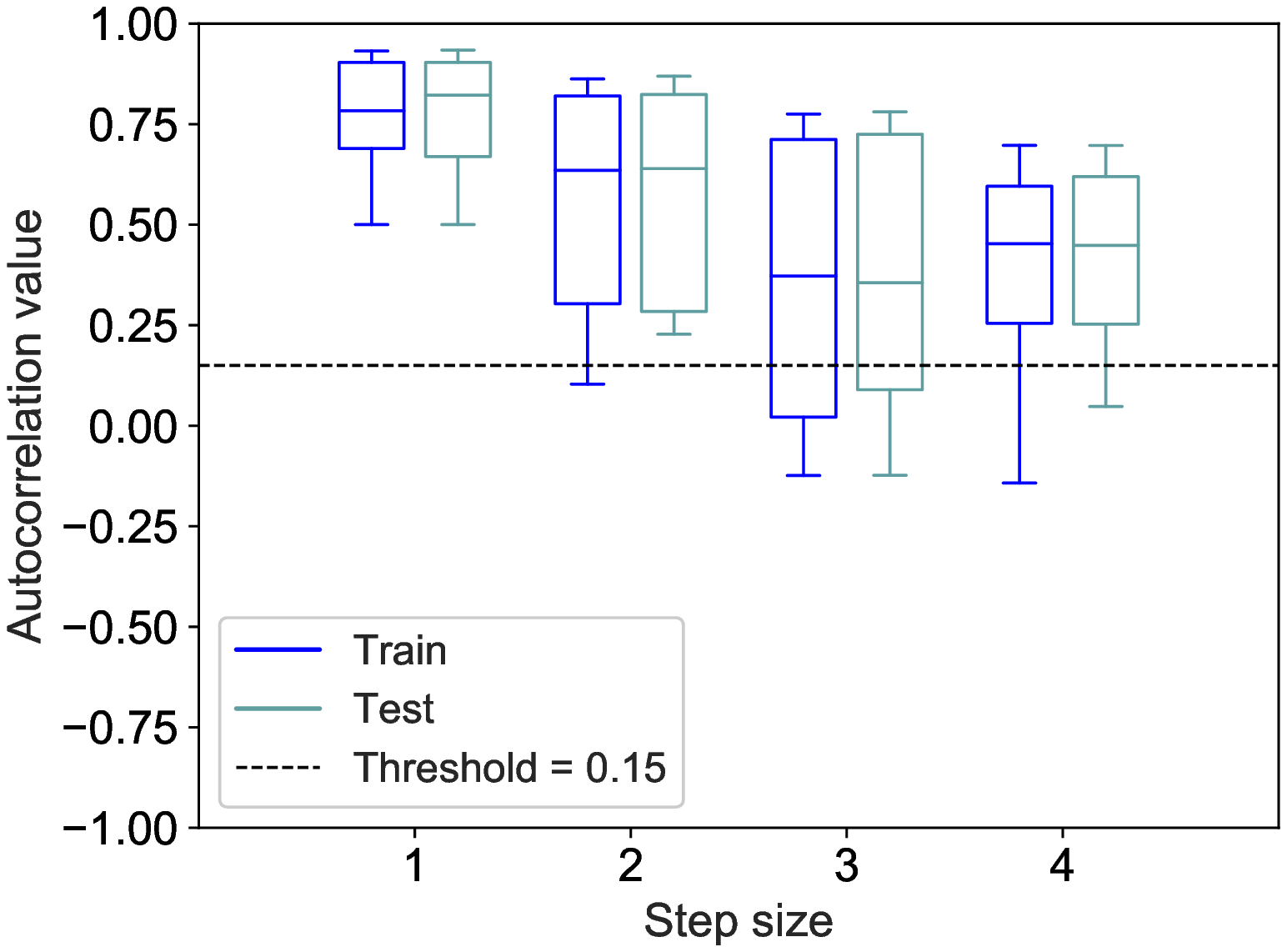}
                \caption{Learning}
                \label{fig:ac-fmnist-learning}
            \end{subfigure}
            \begin{subfigure}{0.31\textwidth}
            \includegraphics[width=\textwidth]{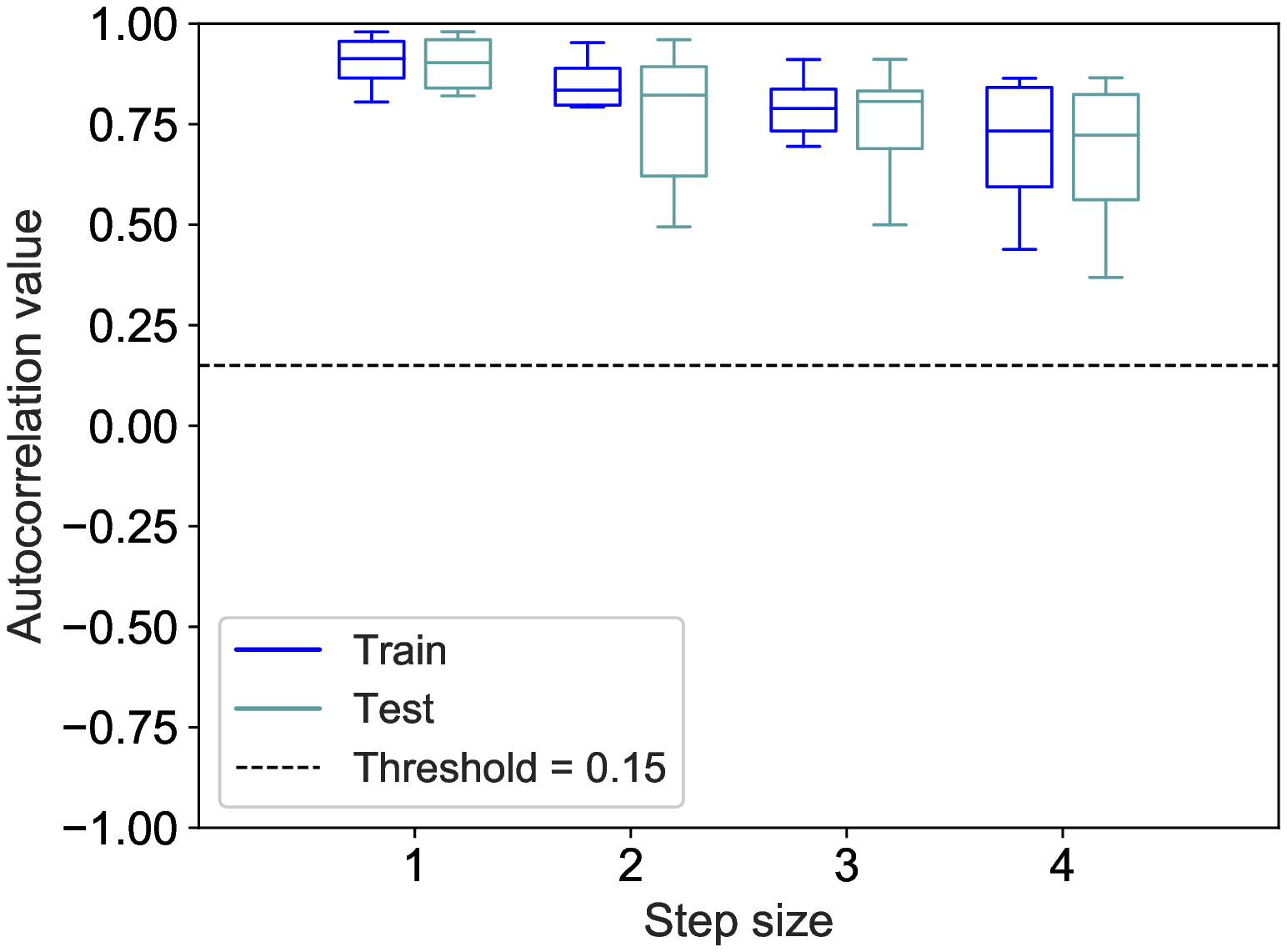}
                \caption{Parameters}
                \label{fig:ac-fmnist-params}
            \end{subfigure}
            \begin{subfigure}{0.31\textwidth}
            \includegraphics[width=\textwidth]{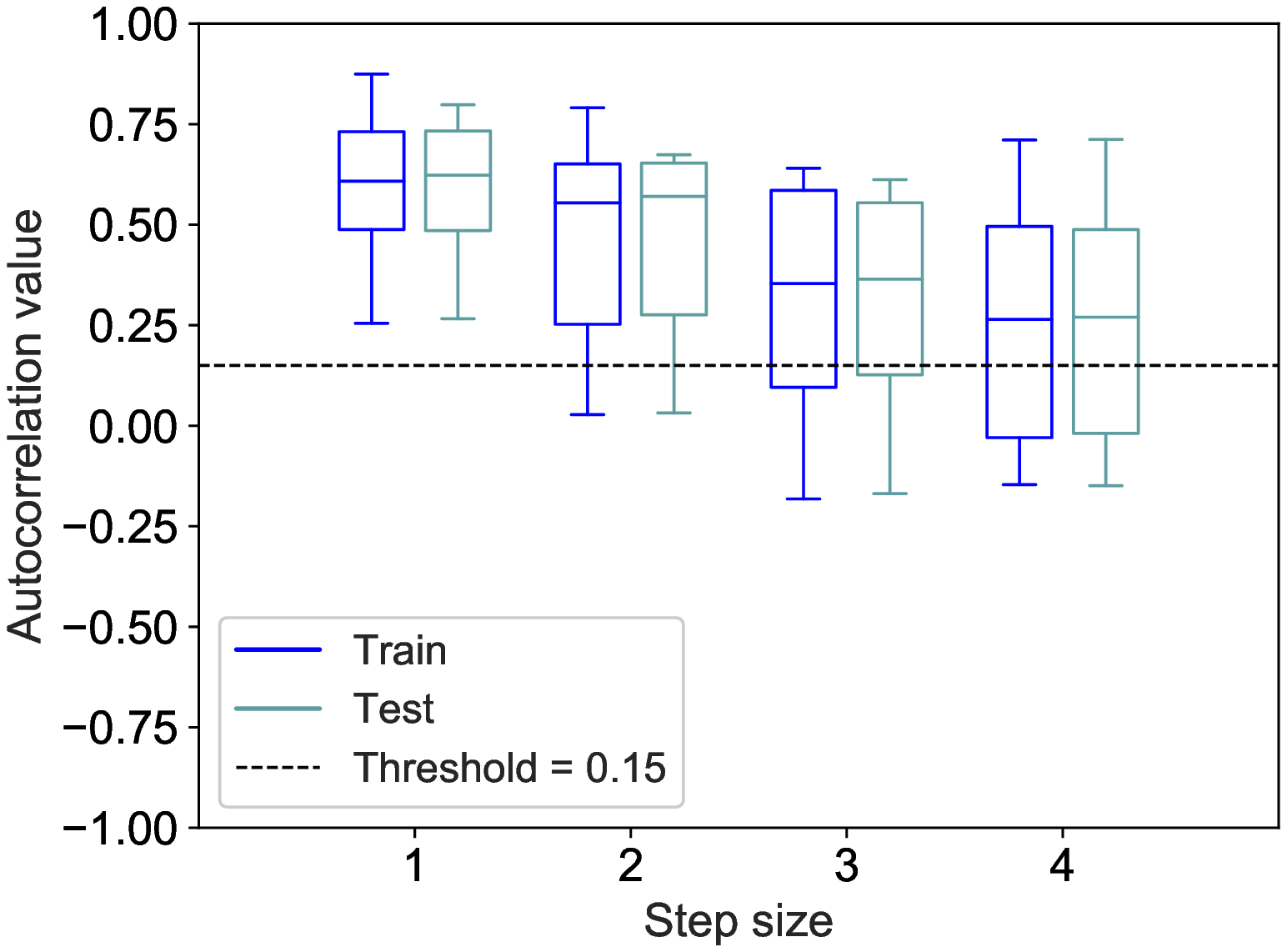}
                \caption{Topology}
                \label{fig:ac-fmnist-topology}
            \end{subfigure}
    \caption{{\it FMNIST dataset}. The organization of the plots is
    analogous to Fig.~\ref{fig:ac-mnist}.}
    \label{fig:ac-fmnist}
\end{figure*}

Now we consider the results obtained for the FMNIST dataset, reported in
Fig.~\ref{fig:ac-fmnist}.
%--------------------------------------------------------------------------------------------
%   
%     BEGIN FIGURE NEUROEVOLUTION VS AUTOCORRELATION
%                                      FMNIST
%
%--------------------------------------------------------------------------------------------
%--------------------------------------------------------------------------------------------
%   
%     END FIGURE NEUROEVOLUTION VS AUTOCORRELATION
%                                      FMNIST
%
%--------------------------------------------------------------------------------------------
%--------------------------------------------------------------------------------------------
%   
%     BEGIN FIGURE NEUROEVOLUTION VS AUTOCORRELATION
%                                      CIFAR10
%
%--------------------------------------------------------------------------------------------
\begin{figure*}[!h]
    \captionsetup[subfigure]{justification=centering}
        \centering
        \begin{subfigure}{0.31\textwidth}
        \hspace*{0.2cm}
            \includegraphics[width=\textwidth]{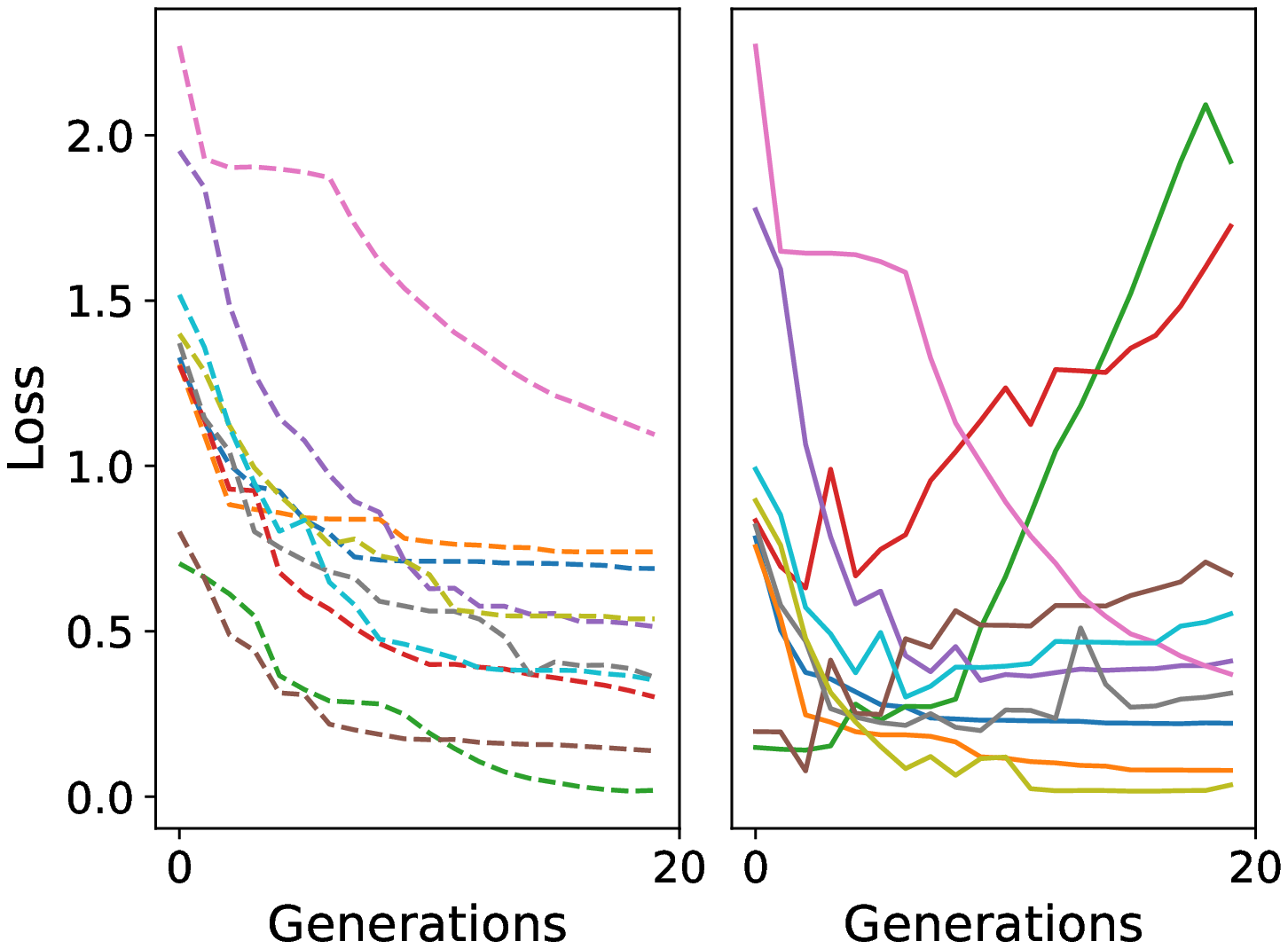}
                \caption{Learning}
                \label{fig:evo-cifar10-learning}
            \end{subfigure}
            \begin{subfigure}{0.31\textwidth}
            \hspace*{0.2cm}
            \includegraphics[width=\textwidth]{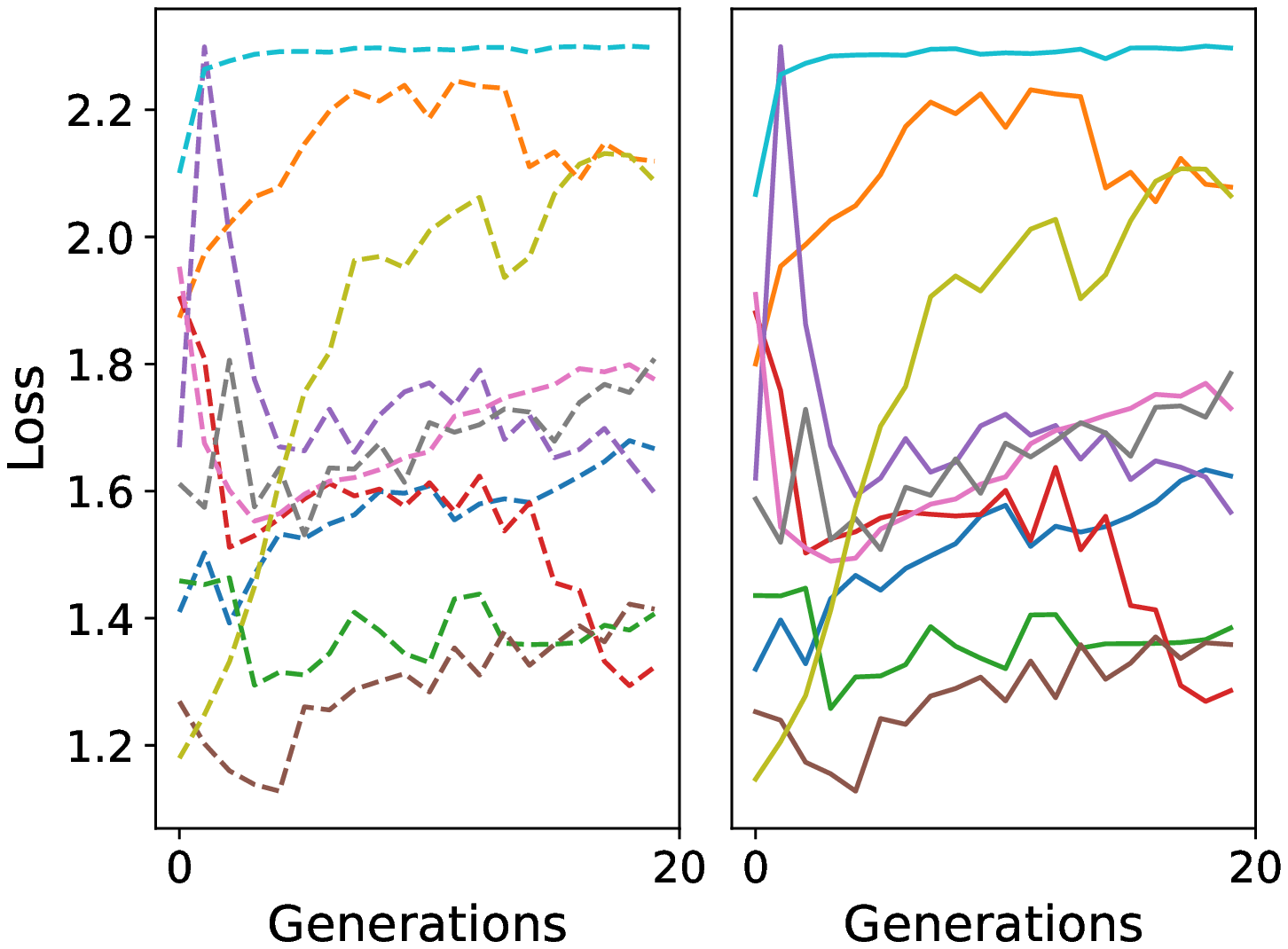}
                \caption{Parameters}
                \label{fig:evo-cifar10-params}
            \end{subfigure}
            \begin{subfigure}{0.31\textwidth}
            \hspace*{0.2cm}
            \includegraphics[width=\textwidth]{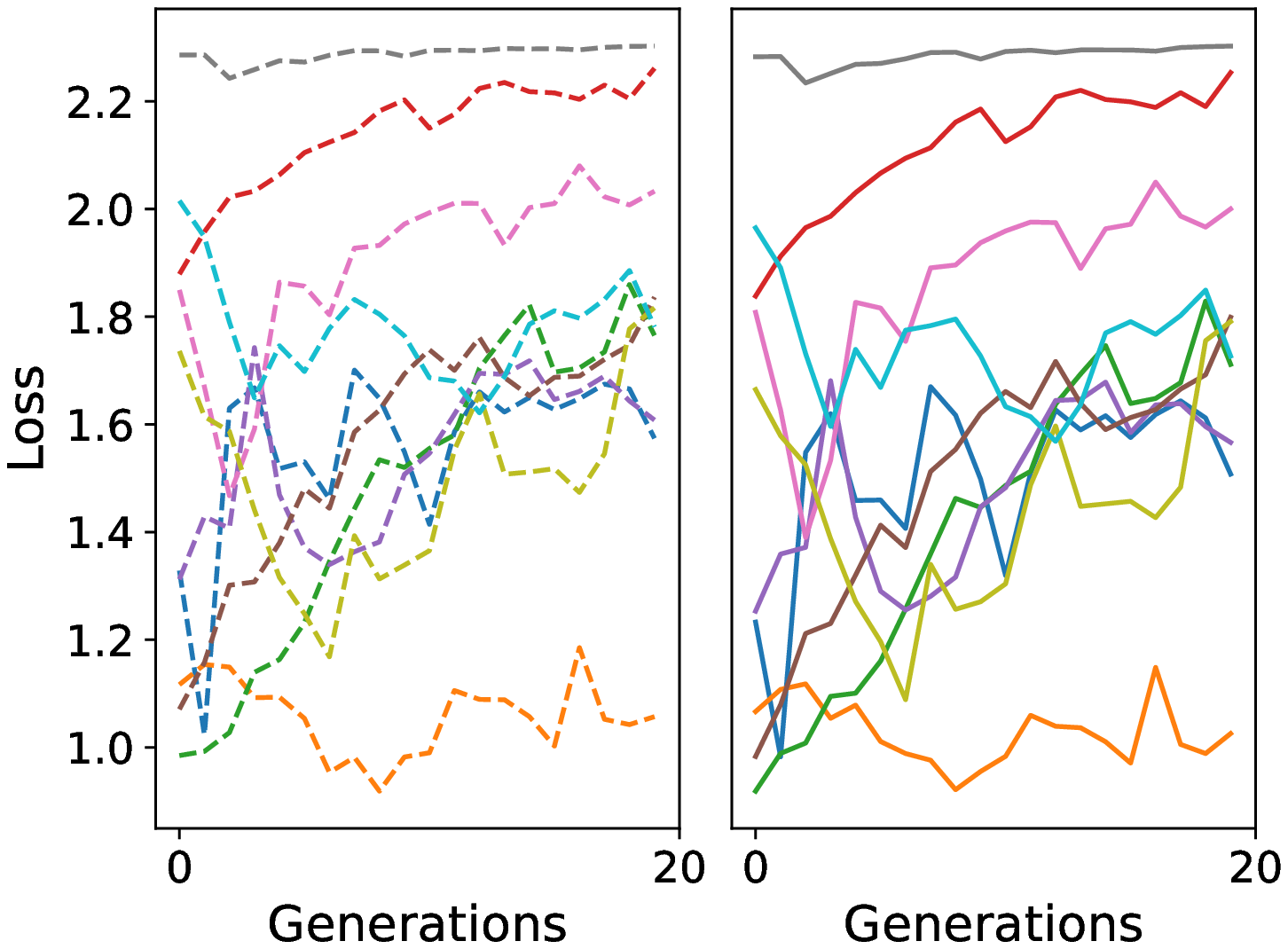}
                \caption{Topology}
                \label{fig:evo-cifar10-topology}
            \end{subfigure}

            \begin{subfigure}{0.31\textwidth}
            \includegraphics[width=\textwidth]{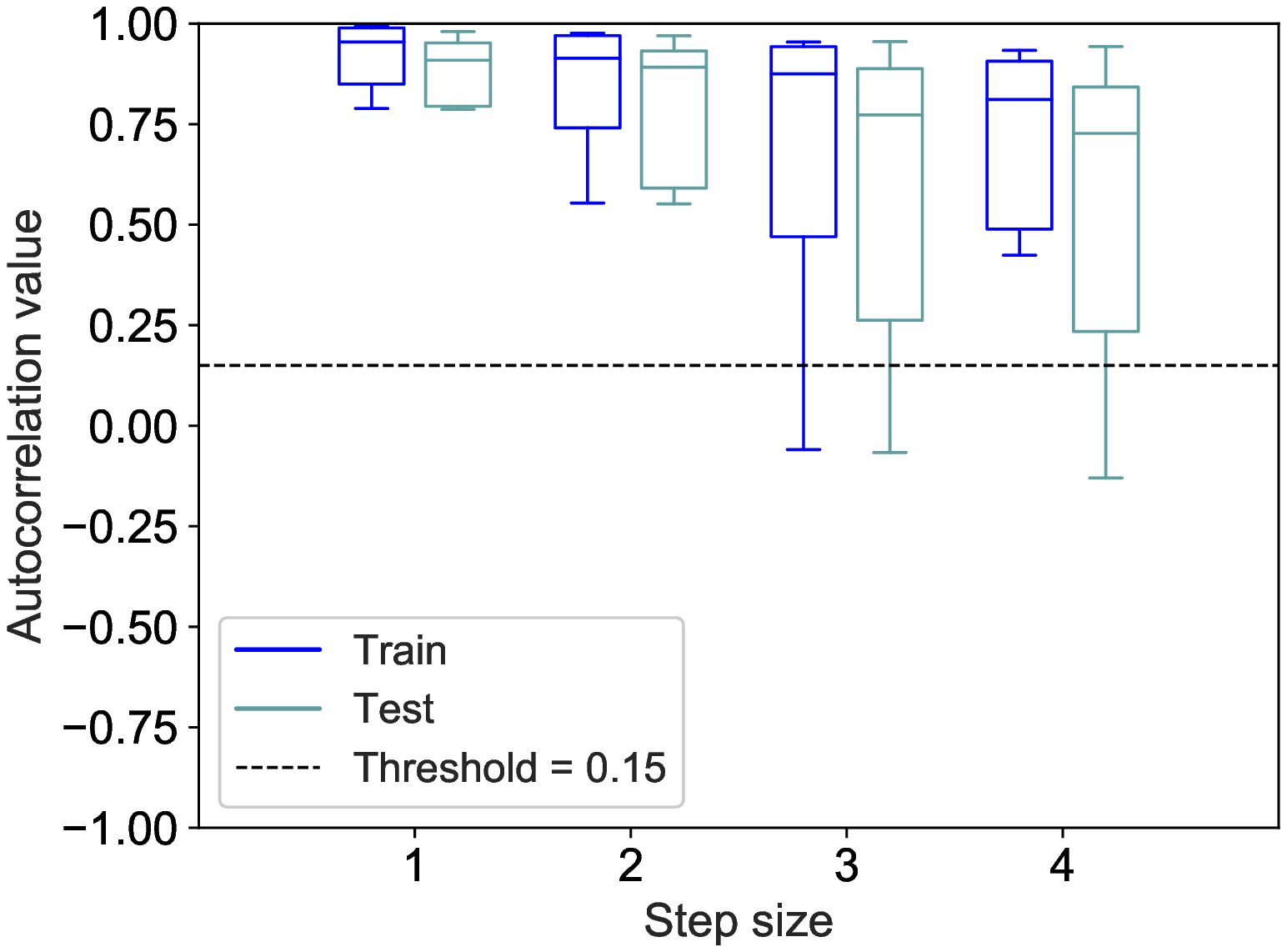}
                \caption{Learning}
                \label{fig:ac-cifar10-learning}
            \end{subfigure}
            \begin{subfigure}{0.31\textwidth}
            \includegraphics[width=\textwidth]{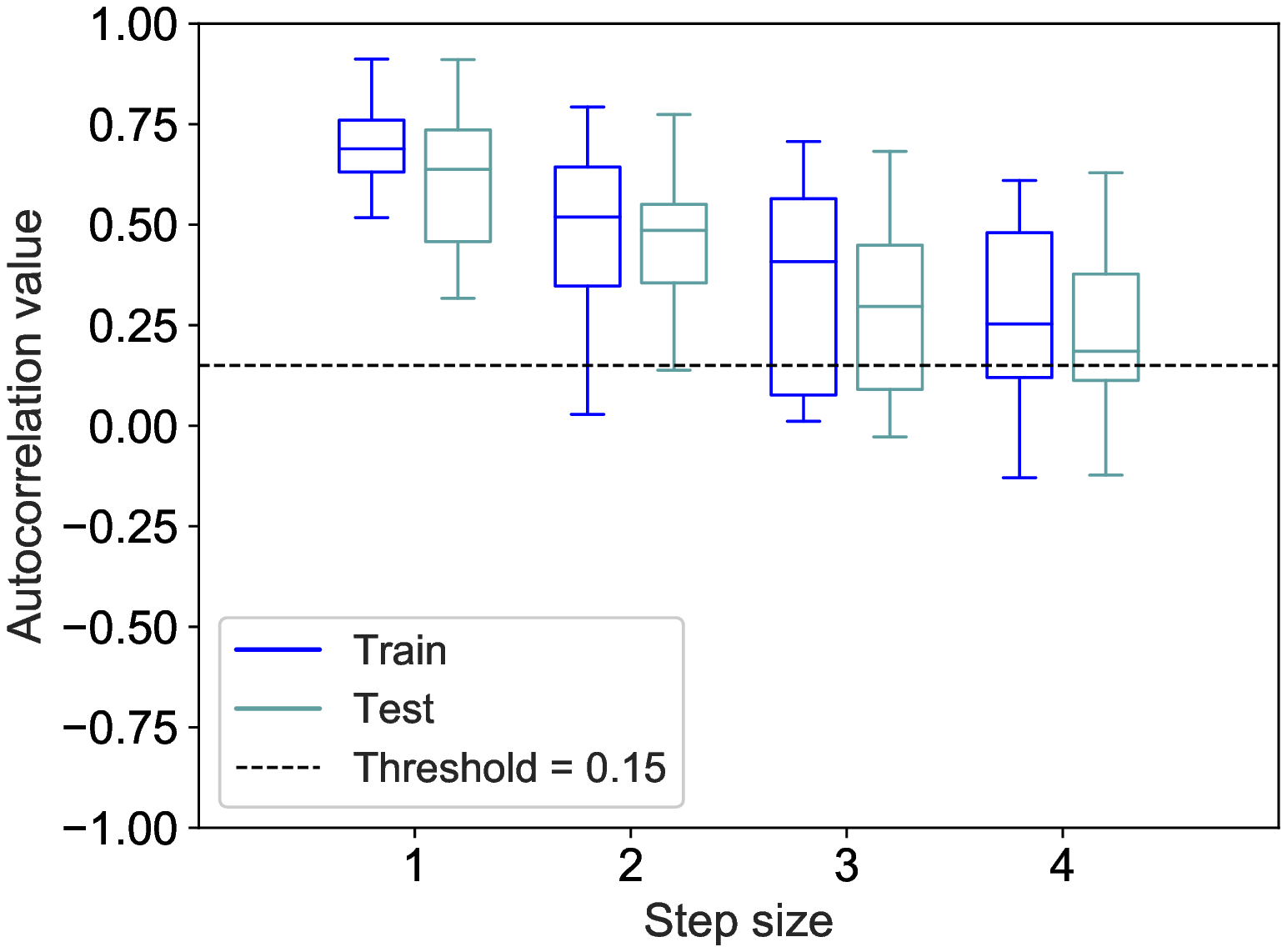}
                \caption{Parameters}
                \label{fig:ac-cifar10-params}
            \end{subfigure}
            \begin{subfigure}{0.31\textwidth}
            \includegraphics[width=\textwidth]{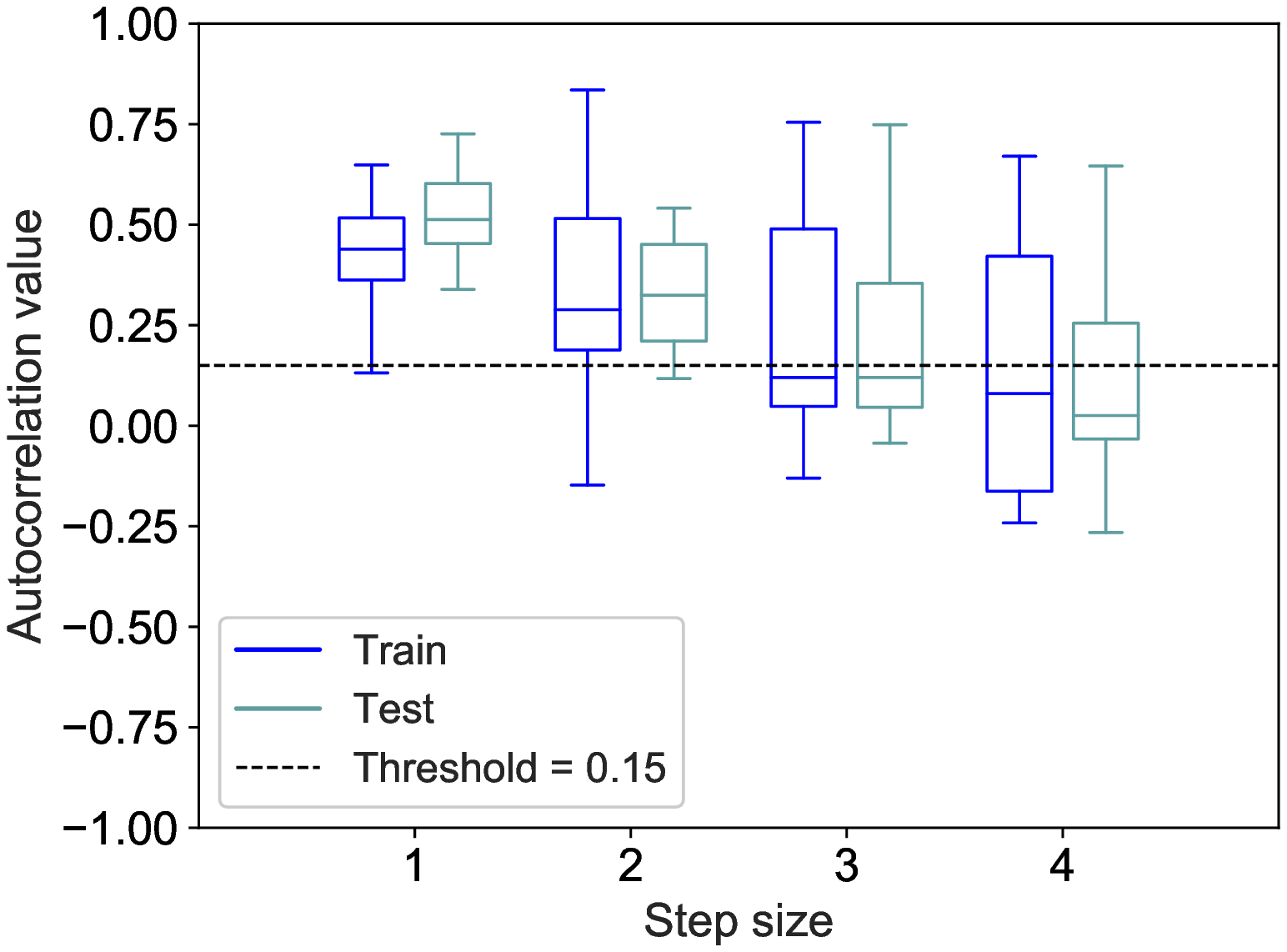}
                \caption{Topology}
                \label{fig:ac-cifar10-topology}
            \end{subfigure}
    \caption{{\it CIFAR10 dataset}. The organization of the plots is
    analogous to Fig.~\ref{fig:ac-mnist}.}
    \label{fig:ac-cifar10}
\end{figure*}%--------------------------------------------------------------------------------------------
%   
%     END FIGURE NEUROEVOLUTION VS AUTOCORRELATION
%                                      CIFAR10
%
%--------------------------------------------------------------------------------------------
%--------------------------------------------------------------------------------------------
%   
%     BEGIN FIGURE NEUROEVOLUTION VS AUTOCORRELATION
%                                      SVHN
%
%--------------------------------------------------------------------------------------------
\begin{figure*}[!h]
    \captionsetup[subfigure]{justification=centering}
        \centering
        \begin{subfigure}{0.31\textwidth}
        \hspace*{0.2cm}
            \includegraphics[width=\textwidth]{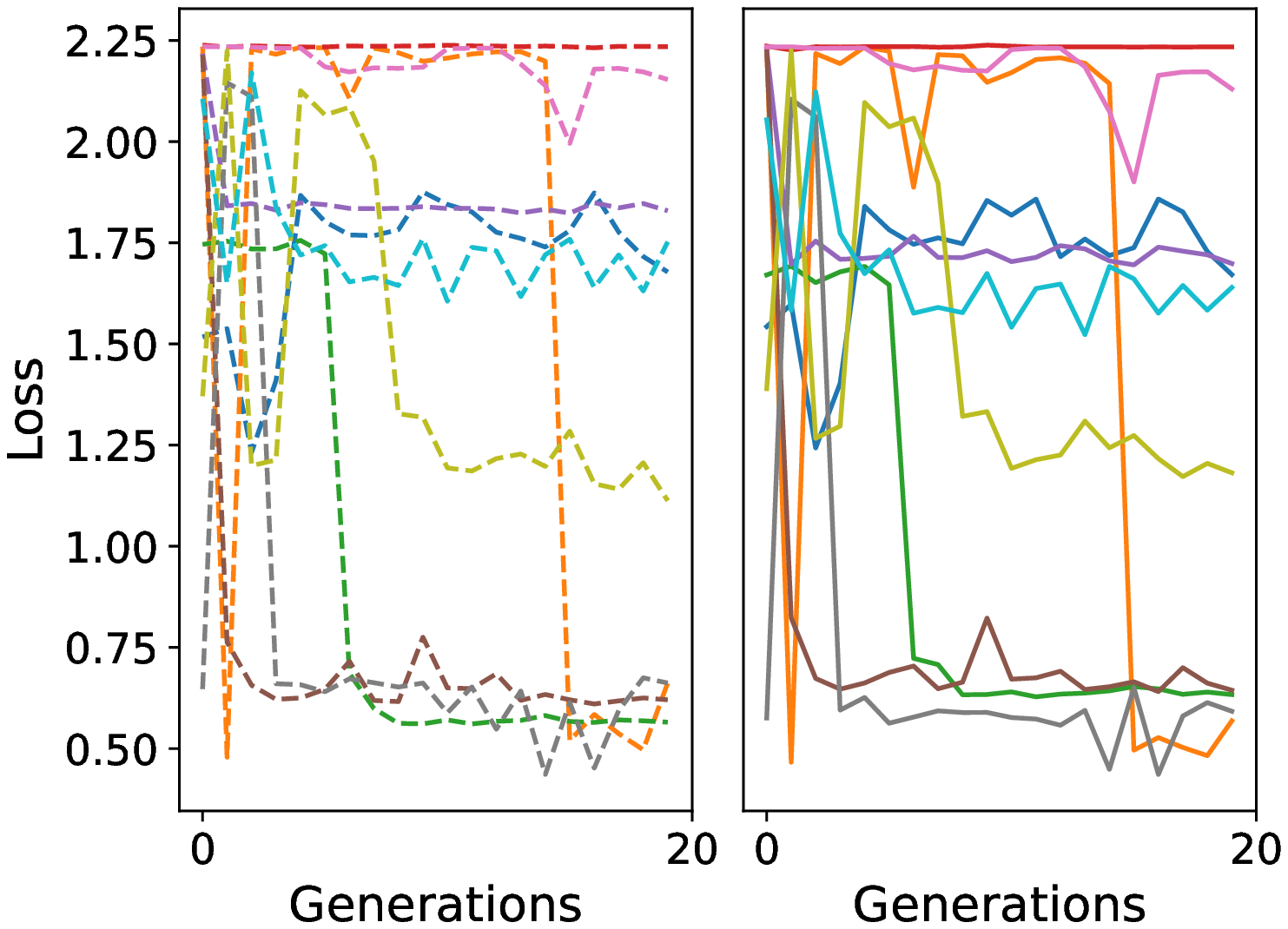}
                \caption{Learning}
                \label{fig:evo-svhn-learning}
            \end{subfigure}
            \begin{subfigure}{0.31\textwidth}
            \hspace*{0.2cm}
            \includegraphics[width=\textwidth]{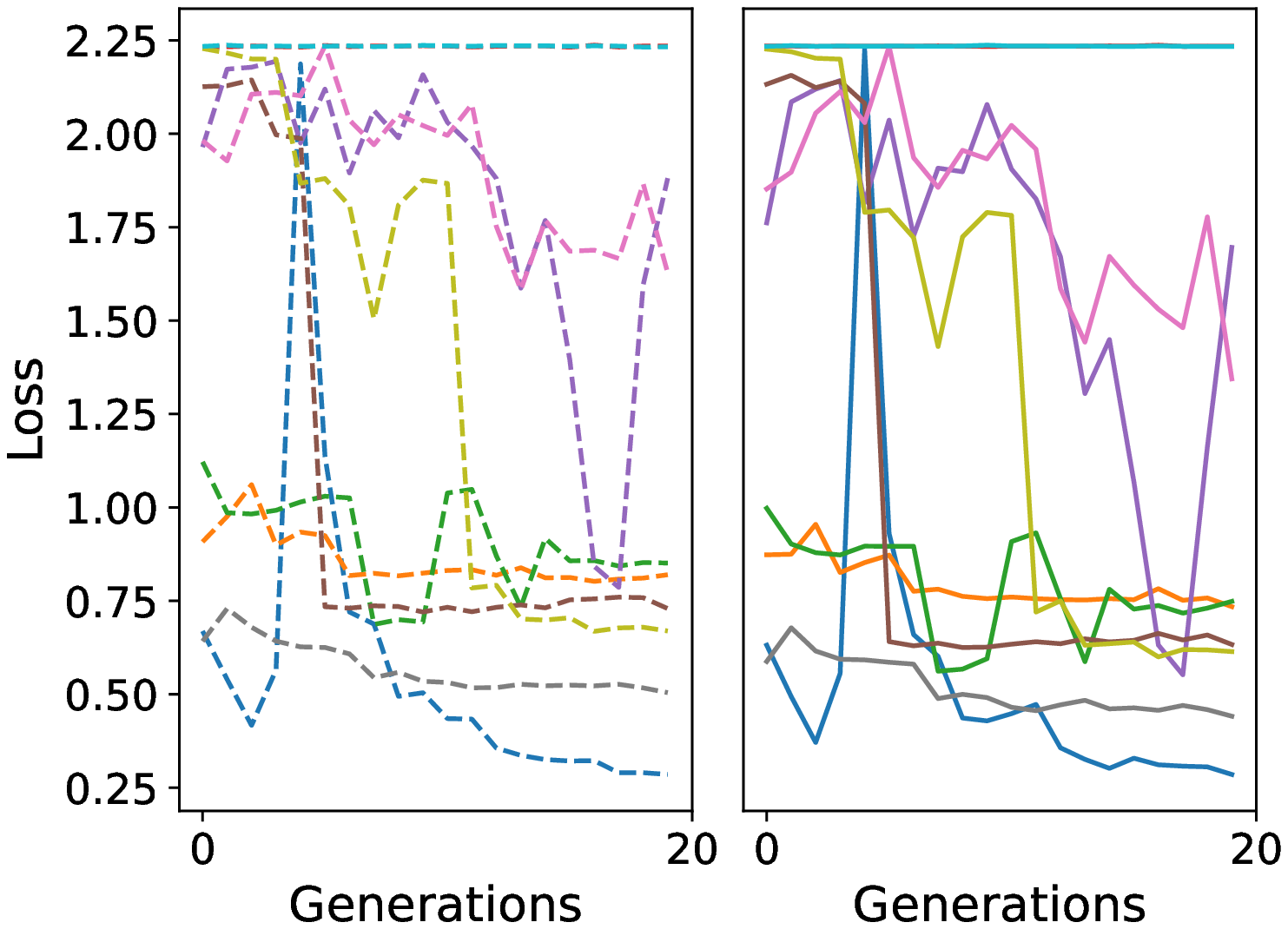}
                \caption{Parameters}
                \label{fig:evo-svhn-params}
            \end{subfigure}
            \begin{subfigure}{0.31\textwidth}
            \hspace*{0.2cm}
            \includegraphics[width=\textwidth]{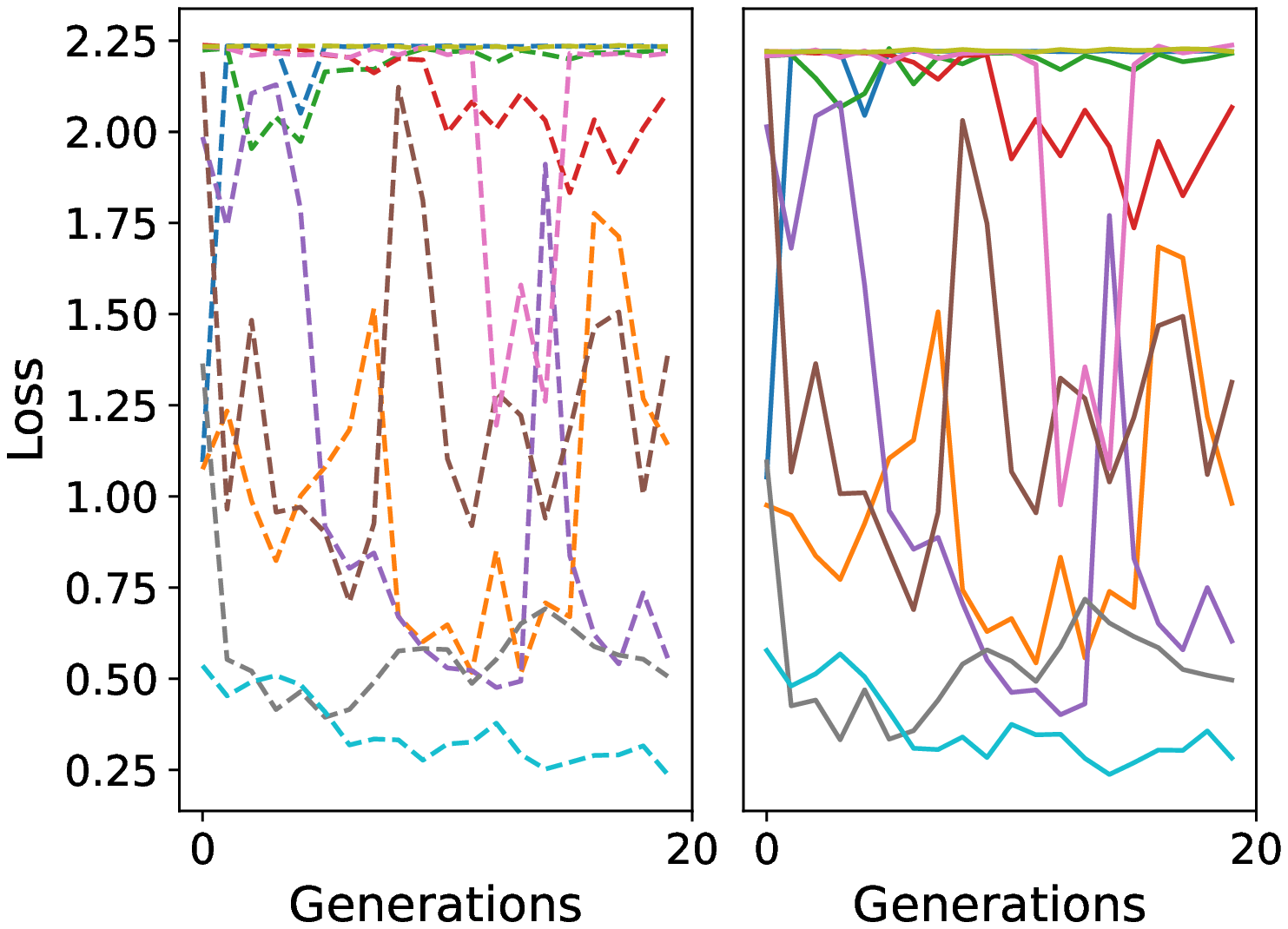}
                \caption{Topology}
                \label{fig:evo-svhn-topology}
            \end{subfigure}

            \begin{subfigure}{0.31\textwidth}
            \includegraphics[width=\textwidth]{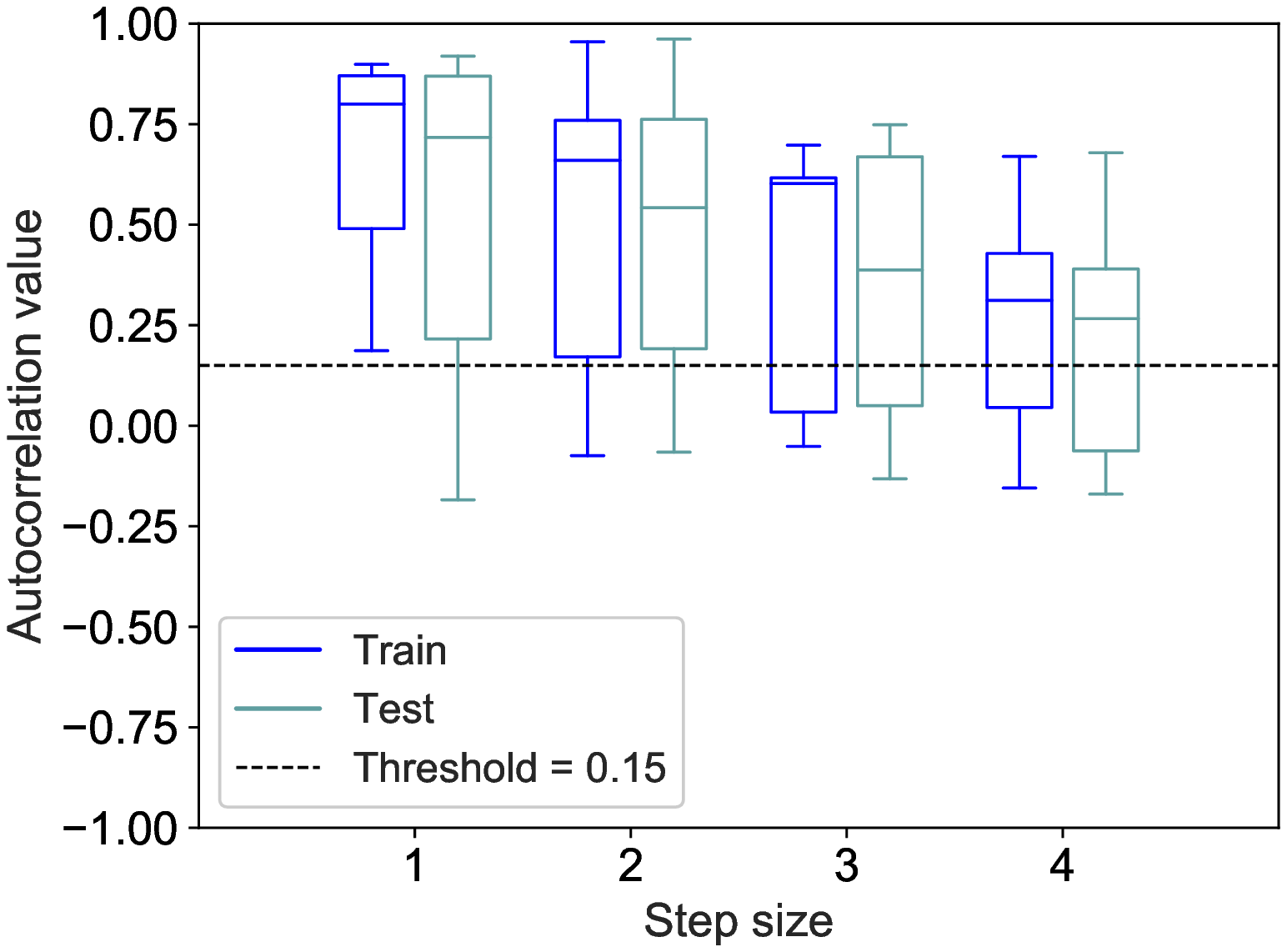}
                \caption{Learning}
                \label{fig:ac-svhn-learning}
            \end{subfigure}
            \begin{subfigure}{0.31\textwidth}
            \includegraphics[width=\textwidth]{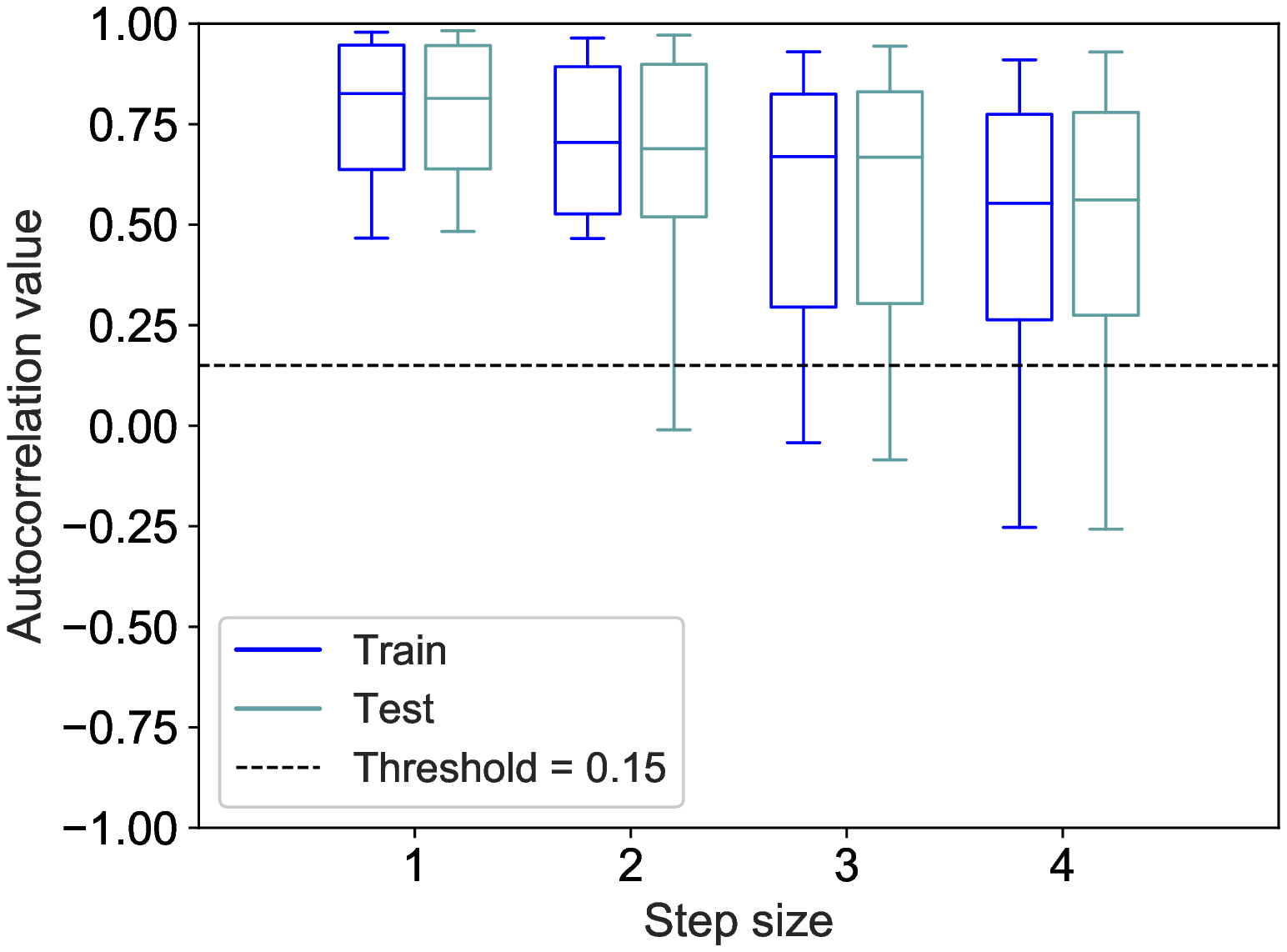}
                \caption{Parameters}
                \label{fig:ac-svhn-params}
            \end{subfigure}
            \begin{subfigure}{0.31\textwidth}
            \includegraphics[width=\textwidth]{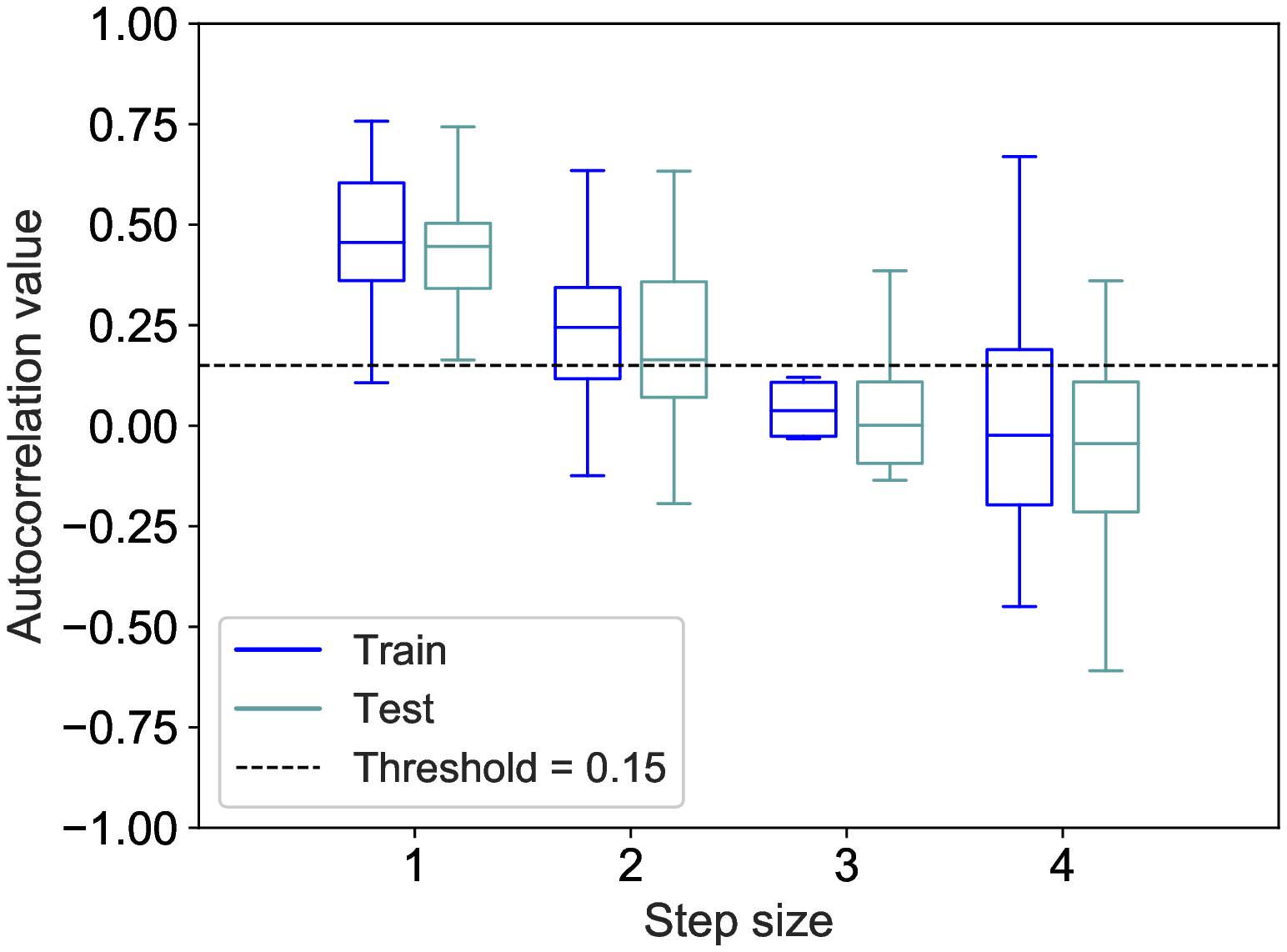}
                \caption{Topology}
                \label{fig:ac-svhn-topology}
            \end{subfigure}
    \caption{{\it SVHN dataset}. The organization of the plots is
    analogous to Fig.~\ref{fig:ac-mnist}.}
    \label{fig:ac-svhn}
\end{figure*}%--------------------------------------------------------------------------------------------
%   
%     END FIGURE NEUROEVOLUTION VS AUTOCORRELATION
%                                      SVHN
%
%--------------------------------------------------------------------------------------------
Describing the results for this dataset is straightforward:
observing the neuroevolution results, we can see that 
for the three configurations the problems are
easy, both on training and test set. In fact, all the curves are steadily
decreasing and/or close to zero.
Observing the scale on the left part of the plots, we can also observe
that when topology mutation is used~(plot~(c)), the problem is slightly harder
than when the other mutations are used, since the achieved values
of the loss are generally higher.
All this is correctly predicted
by the autocorrelation, given that the boxes are above the threshold
for all the configurations,
and, in the case of the topology mutation~(plot~(f)), they are slightly lower
than in the other cases. 
Last but not least, also in this case training and test evolution of loss are very
similar between each other, and this fact finds a precise correspondence in the
autocorrelation results, given that the training boxes are generally very similar
to the test boxes.
All in all, we can conclude that also for the FMNIST dataset, autocorrelation
is a reliable indicator of problem hardness.

The results for the CIFAR10 dataset are reported in Fig.~\ref{fig:ac-cifar10}.
Observing the neuroevolution results, we can say that when the learning mutation
is used, the problem is easy (almost all the loss curves have a smooth decreasing trend,
except for some outliers on the test), when the parameters mutation is used the problem
is uncertain (in some runs the loss curves have a decreasing trend,
while in others they have an increasing trend), and when the topology mutation is used, the problem is hard (almost all the loss curves have an increasing trend).
Observing the autocorrelation results, we find a reasonable 
correspondence. For the learning mutation all the boxes are clearly
above the threshold, for the parameters mutation the boxes are not as high as for the learning mutation, beginning to cross the threshold with steps~3 and~4, and finally for the topology mutation
the boxes are even lower, with the medians below the threshold for steps~3 and~4, and more than half the height of the boxes also below the threshold for step~4.
As already observed in plot~(f) of Fig.~\ref{fig:ac-fmnist}, longer steps
seem to be better indicators when the autocorrelation is applied to hard problems.
Another observation for CIFAR10 is that the evolution of the loss curves for the learning mutation (plot~(a)) clearly shows that there is a much larger diversity of behaviors on the test set than on the training set. This fact also finds a correspondence in the
autocorrelation results (plot~(d)), given that the test boxes are taller than the
training boxes, in particular for step~4.

As a last test case for autocorrelation, we now analyse the results obtained on the SVHN dataset, reported
in Fig.~\ref{fig:ac-svhn}.
In this case, the plots of the loss evolution indicate that the problem
is uncertain when learning mutation is used (given that approximately half
of the curves have a decreasing trend, while the other half have an oscillating
trend), easy when parameters mutation is used (with the majority of the
curves having a decreasing trend) and hard
when topology mutation is used (with most curves exhibiting an oscillatory behaviour, which indicates poor optimization ability).
Also in this case, autocorrelation is a reasonable indicator of problem
difficulty. For learning mutation the boxes are crossing the threshold for steps~3 and~4, for parameters mutation they are above the threshold, and for topology mutation they are almost completely below the threshold for steps~3 and~4.
The medians are lower and the dispersion of values is larger for step~4, which reflects well the neuroevolution behavior observed in plot~(c) (unstable and often returning to high values of the loss). The highest step length is once again the most reliable.

%--------------------------------------------------------------------------------------------
%   
%     BEGIN FIGURE ENTROPY MEASURE OF RUGGEDNESS
%
%--------------------------------------------------------------------------------------------
\begin{figure*}[!h]
    \captionsetup[subfigure]{justification=centering}
        \centering
          \begin{subfigure}{0.31\textwidth}
            \includegraphics[width=\textwidth]{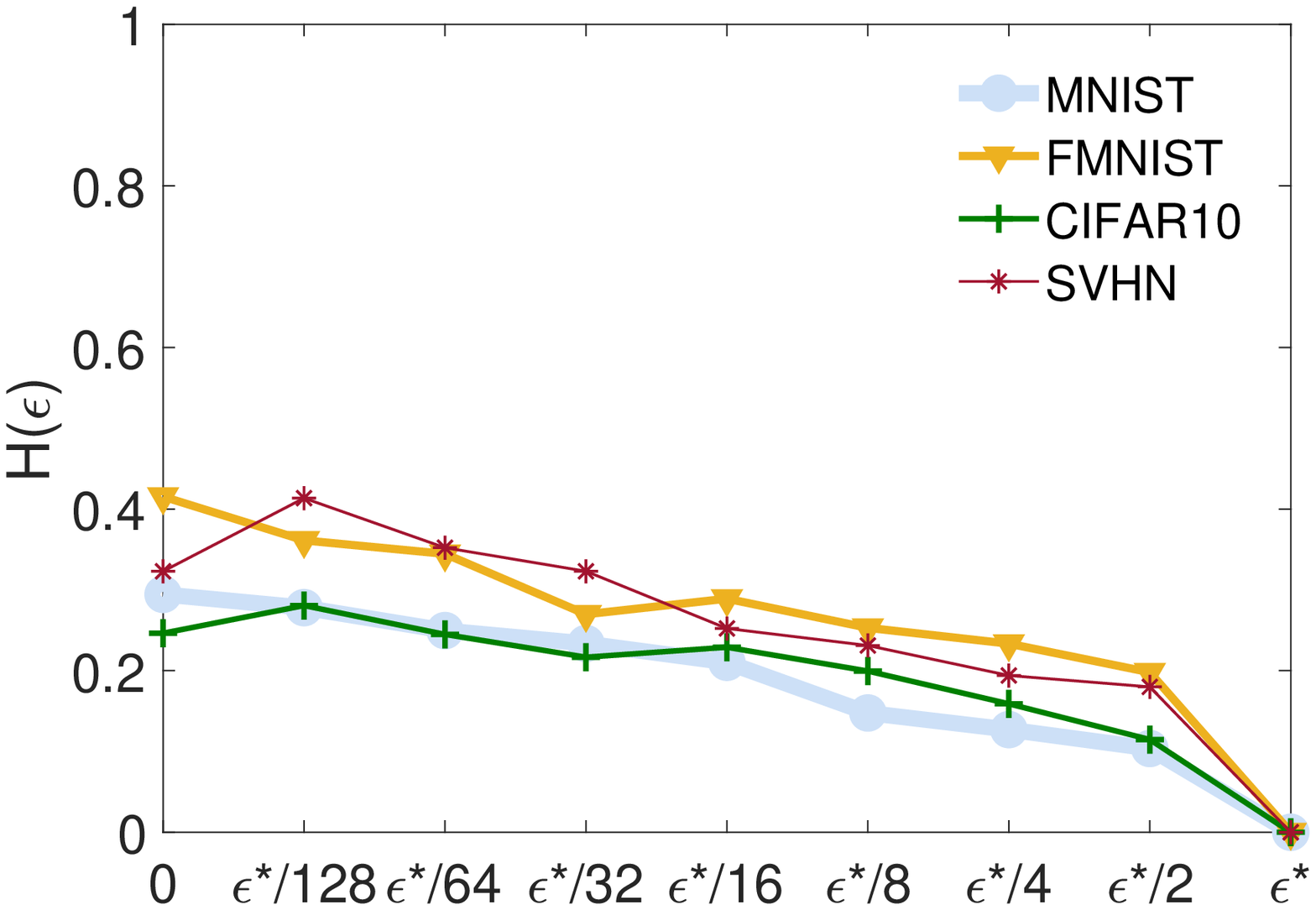}
              \caption{Learning}
              \label{fig:entropy-learning}
          \end{subfigure}
          \begin{subfigure}{0.31\textwidth}
            \includegraphics[width=\textwidth]{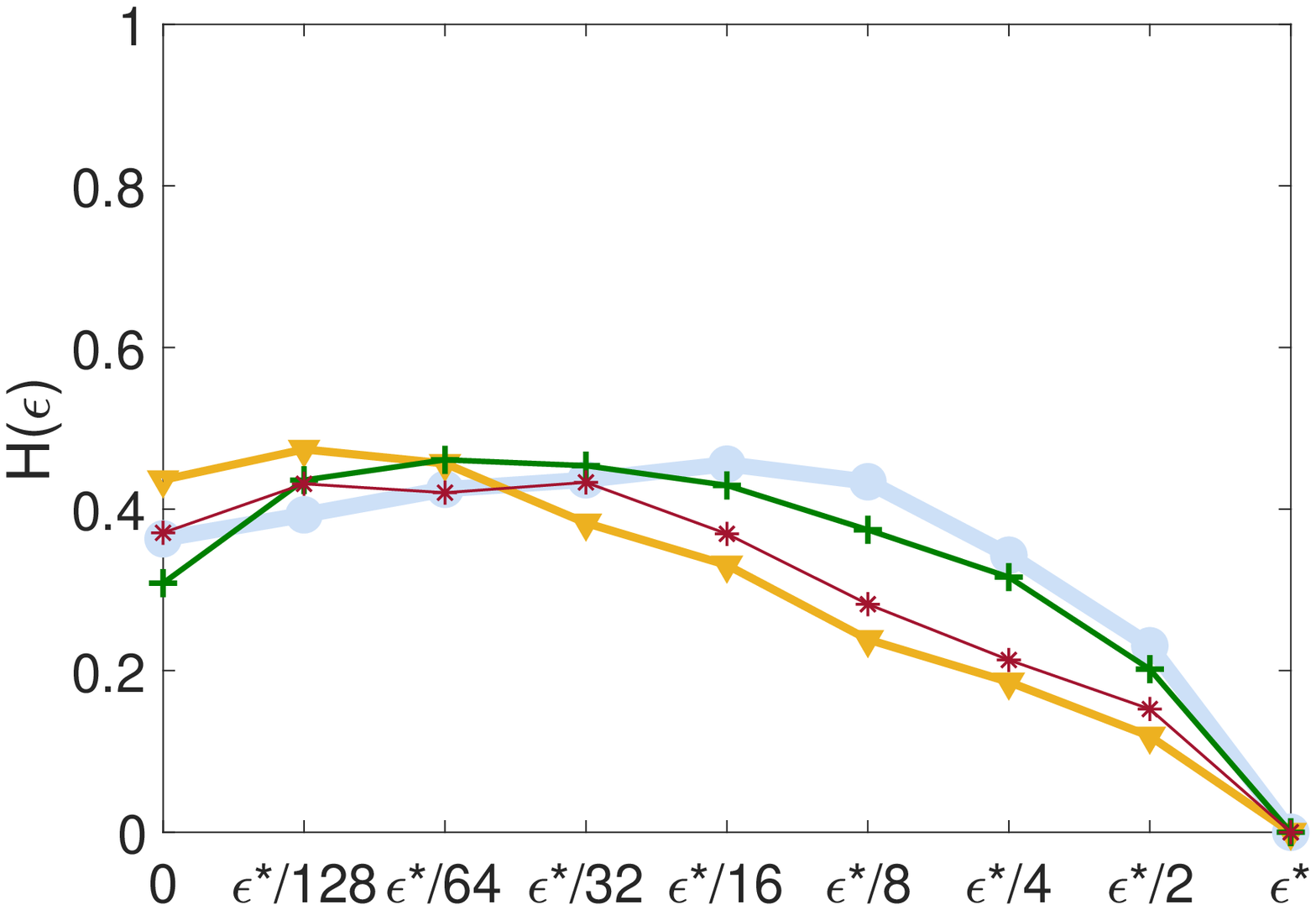}
              \caption{Parameters}
              \label{fig:entropy-params}
          \end{subfigure}
          \begin{subfigure}{0.31\textwidth}
            \includegraphics[width=\textwidth]{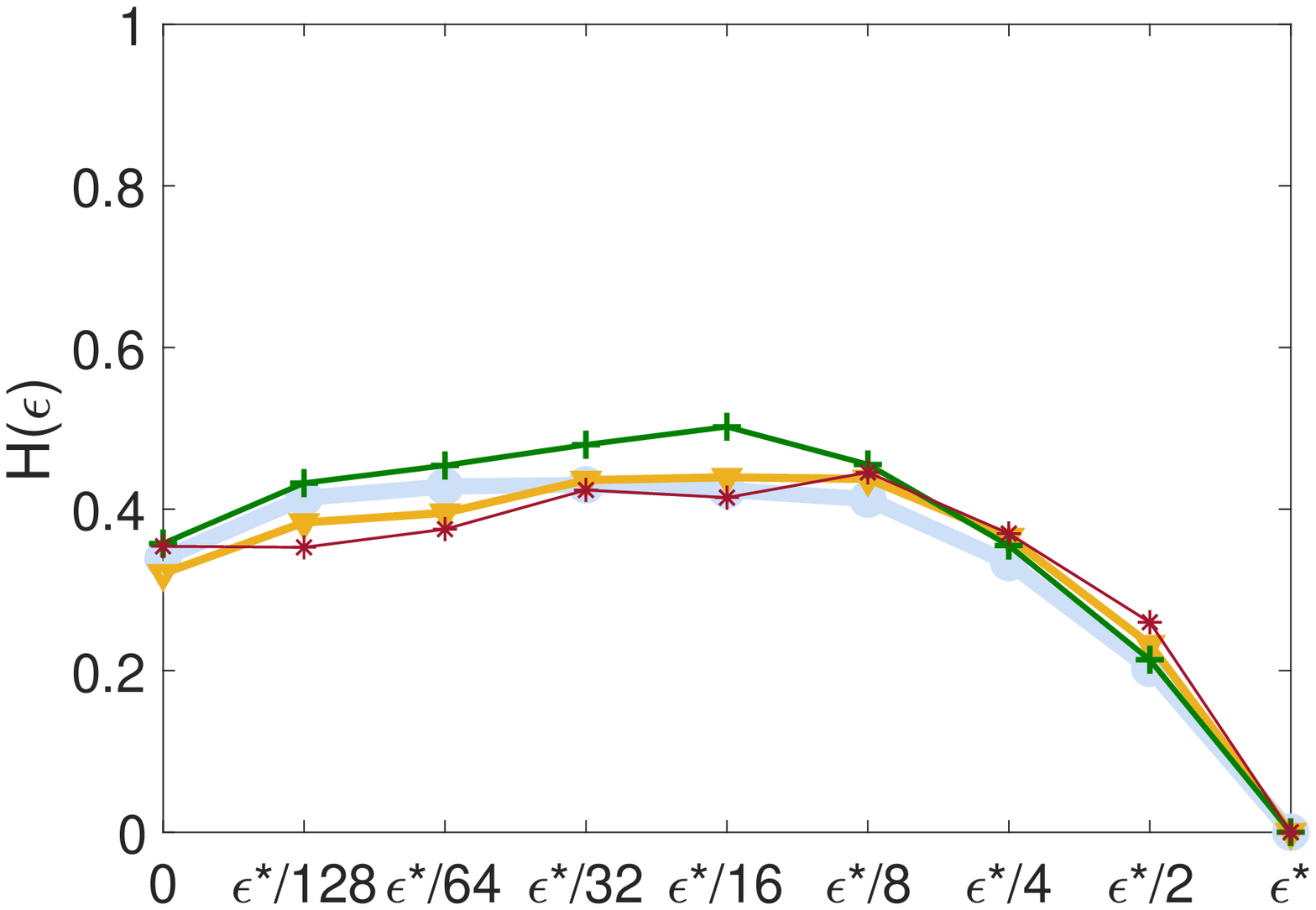}
              \caption{Topology}
              \label{fig:entropy-topology}
          \end{subfigure}
    \caption{Results of the Entropic Measure of Ruggedness $\bar{H}(\varepsilon)$  over different values of $\varepsilon^*$ for the the three mutation operators on the four considered test problems.}
    \label{fig:entropy}
    \vspace{-6pt}
\end{figure*}
%--------------------------------------------------------------------------------------------
%   
%     END FIGURE ENTROPY MEASURE OF RUGGEDNESS
%
%--------------------------------------------------------------------------------------------

%------------------------
%
%  BEGIN TABLE
%  EMR Results
%
%------------------------
\begin{table}[b]
\caption{$R_{f}$ for each mutation on the studied test problems.}
\label{table:rf}
\centering
\begin{tabular}{c|ccc}
        & \textbf{Learning} & \textbf{Parameters} & \textbf{Topology} \\ \hline
MNIST   & 0.29     & 0.45       & 0.43     \\ %\hline
FMNIST  & 0.41     & 0.47       & 0.43     \\ %\hline
CIFAR10 & 0.28     & 0.46       & 0.50     \\ %\hline
SVHN    & 0.41     & 0.43       & 0.44    
\end{tabular}
\end{table} 
%------------------------
%
%  END TABLE
%  EMR Results
%
%------------------------
Finally, we study the results of the~EMR, reported in Fig.~\ref{fig:entropy}.
Each plot reports the results for one mutation type, showing the values of $H(\varepsilon)$ for multiple $\varepsilon$ values (see Sect.~\ref{entrmeasofrug})
%\mbox{$\varepsilon = \{0, \frac{\varepsilon^*}{128}, \frac{\varepsilon^*}{64}, \frac{\varepsilon^*}{32}, \frac{\varepsilon^*}{16}, \frac{\varepsilon^*}{8}, \frac{\varepsilon^*}{4}, \frac{\varepsilon^*}{2}, \varepsilon^* \}$}
on the 
four studied datasets.
These curves illustrate the trend of how ruggedness changes with respect to neutrality.
The results show that, overall, the obtained landscapes have a low degree of neutrality, not maintaining the value of~$H(\varepsilon)$ as~$\varepsilon$ increases.
The most neutral landscape is the one produced by topology mutation on the MNIST dataset (plot~(c) of Fig.~\ref{fig:entropy}).
Its highest $H(\varepsilon)$ happens when $\varepsilon = \varepsilon^*/64$, but the value suffers minimal change from $\varepsilon = \varepsilon^*/128$ to
 $\varepsilon = \varepsilon^*/8$.

Table~\ref{table:rf}, which reports the values of~$R_f$ for each type of mutation, and for each studied test problem, corroborates the previous discussion: the maximum value for learning mutation is~0.41, while for parameters mutation is~0.47 and for topology mutation is~0.5. 
 Again, we can see that learning mutations induce the smoothest landscapes, while topology mutations induce the most rugged ones.
Also in this case, the prediction of the~EMR corresponds to what we can
observe from the actual neuroevolution runs.

%%%%%%%%%%%%%%%%%%%%%%%%%%%%%%%%
%
%   Conclusions
%
%%%%%%%%%%%%%%%%%%%%%%%%%%%%%%%%
\section{Conclusions and Future Work}
\label{concl}

Two different measures of fitness landscapes, autocorrelation and
entropic measure of ruggedness, were used for the first time 
to characterize the performance of neuroevolution of convolutional
neural networks.
The results were obtained on four different test problems, and confirm that
these measures are reasonable indicators of problem hardness,
both on the training set and on the test set.
Future work involves the study of other measures of fitness landscapes,
on more test problems, with the objective of developing well established,
theoretically motivated predictive tools for neuroevolution, that can
significantly simplify the configuration and tuning phases of the algorithm.
Being able to use more powerful computational architectures,
so that we are able to calculate the measures on larger and more
significant samples of solutions, is crucial for achieving such an
ambitious goal.
~\\

%%%%%%%%%%%%%%%%%%%%%%%%%%%%%%%
%
%   Acknowledgments
%
%%%%%%%%%%%%%%%%%%%%%%%%%%%%%%%%
\section*{Acknowledgments}
This work was partially supported by FCT through funding of LASIGE Research Unit (UIDB/00408/2020) and projects PTDC/CCI-INF/29168/2017, PTDC/CTA-AMB/30056/2017, PTDC/CCI-CIF/29877/2017, PTDC/ASP-PLA/28726/2017, DSAIPA/DS/0022/2018, DSAIPA/DS/0113/2019.

%%%%%%%%%%%%%%%%%%%%%%%%%%%%%%%
%
%   Bibliography
%
%%%%%%%%%%%%%%%%%%%%%%%%%%%%%%%
\bibliographystyle{IEEEtran}
\bibliography{bibliography}

%%%%%%%%%%%%%%%%%%%%%%%%%%%%%%%
%
%   End Document
%
%%%%%%%%%%%%%%%%%%%%%%%%%%%%%%%
\end{document}